\documentclass{article}
\usepackage[margin=1in]{geometry}
 
\usepackage[utf8]{inputenc}
\usepackage[T1]{fontenc}
\usepackage[table, svgnames, dvipsnames]{xcolor}

\definecolor{RoyalBlue}{HTML}{ED1B23}

\usepackage{graphicx}
\usepackage{hyperref}
\usepackage{booktabs}
\usepackage{amsfonts}
\usepackage{nicefrac}
\usepackage{microtype}
\usepackage{float}
\usepackage{amsmath, amssymb, amsthm, mathtools}
\usepackage{multirow}
\usepackage{tabularx}
\usepackage{array}
\usepackage{colortbl}
\usepackage{siunitx}
\usepackage[ruled,vlined]{algorithm2e}
\usepackage{caption}
\usepackage[authoryear]{natbib}
\usepackage{wrapfig}

\hypersetup{
    colorlinks=true,
    linkcolor=Red,
    citecolor=blue,
    filecolor=magenta,
    urlcolor=Red,
}

\emergencystretch=\maxdimen
\numberwithin{equation}{section}
\allowdisplaybreaks

\makeatletter
\newcommand{\vast}{\bBigg@{3}}
\newcommand{\Vast}{\bBigg@{4}}
\makeatother

\newcommand{\EX}{\mathbb{E}}

\renewcommand{\text}[1]{\textnormal{#1}}

\newcommand*{\belowrulesepcolor}[1]{\noalign{\kern-\belowrulesep\begingroup\color{#1}\hrule height\belowrulesep\endgroup}}
\newcommand*{\aboverulesepcolor}[1]{\noalign{\begingroup\color{#1}\hrule height\aboverulesep\endgroup\kern-\aboverulesep}}

\title{Shortcomings and capacities of real-constrained neural networks in complex spaces}

\author{Andrew Gracyk \\
      Department of Mathematics\\
      Purdue University\\
      West Lafayette, IN 47907, United States
      \\
      \texttt{agracyk@purdue.edu}}

\begin{document}

\date{}

\maketitle

\begin{abstract}
\noindent We find the asymptotic ratio between the storage capacities when enforcing real pre-activations in a complex hypothesis class as opposed to complex ones in the same class. We use weights drawn from the complex Gaussian, which converge asymptotically in norm to the square root of dimension almost surely. Our methods depend on Gardner volume-type comparisons at critical capacity. Our proof relies on an application of the Harish-Chandra-Itzykson-Zuber (HCIZ) formula, nonstandard in literature. With the HCIZ formula, we may obtain a more robust approximation for the final asymptotic ratio. This strategy is applicable to our work specifically since we integrate over the unitary and orthogonal compact manifolds, facilitated via the Weyl integration formula and the Haar measure. 
\end{abstract}

\vspace{2mm}


\vspace{2mm}

\noindent\textbf{AMS MSC Classifications (2020):}  	82B44, 60B20, 82C32, 05E05

\tableofcontents

\begin{figure}[!t]
  \centering
  \includegraphics[width=0.6\linewidth]{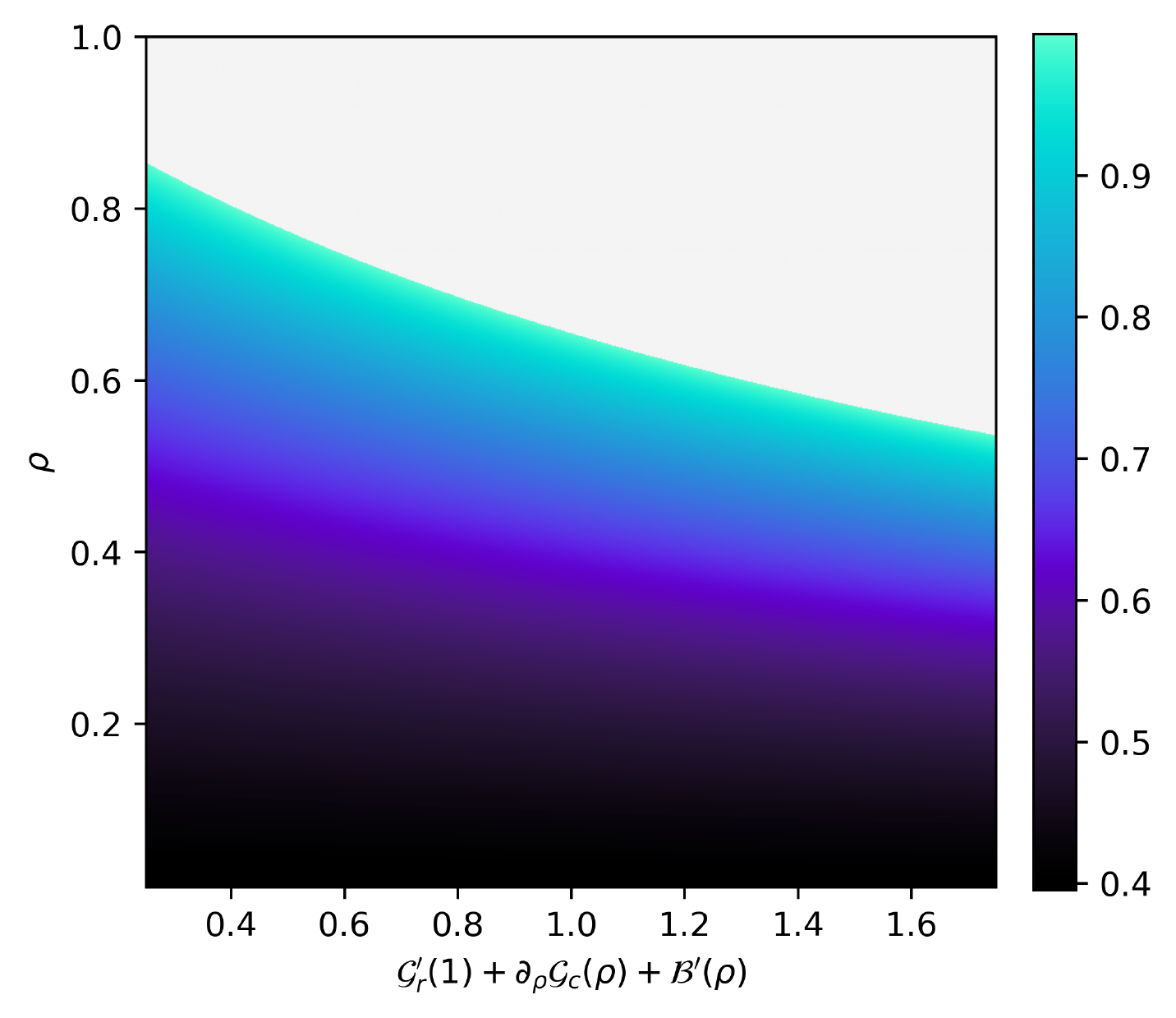}
  \vspace{-2mm}
  \caption{We plot $\Gamma$ as in \ref{sec:main_result} across varying $\rho$ and $\mathcal{G}_{\{r,c\}},\mathcal{B}$. We suppress values of $\Gamma \notin [0,1]$ in the plot. In theory, $\mathcal{G}_{\{r,c\}},\mathcal{B}$ are not free parameters, and $\Gamma \notin [0,1]$ is not an actual scenario. We take $X$ to be plot value and $\mathcal{B}' = X - \mathcal{G}_r' - \partial_\rho \mathcal{G}_c$. We create this plot using a real one-layer complex neural network with normalized weights, and we construct $\mathcal{G}_{r}', \partial_\rho \mathcal{G}_c, \mathcal{B}'$ systematically across a perturbed range using a base value generated from the real network. We choose network width 2000, 3 network copies, with $\kappa=2$. Our values for $\mathcal{G}_r, \mathcal{G}_c$ imitate what we see in section \ref{sec:main_result} with a network but under a special case $q_{\text{empirical}} < 1$ in order to evaluate at the pole. The void space is plausible mathematically but does not represent a realistic scenario in regard to physical constraints on the spin glass system, which we elaborate on in section \ref{sec:discussion}. }
  \label{fig:gamma}
\end{figure}

\section{Introduction}
\label{sec:introduction}

We attempt to usher in new perspectives to machine learning under settings of complex spaces. Complex manifolds and Banach spaces are largely underdeveloped in modern artificial intelligence, although earlier breakthroughs utilized complex data to achieve results new in literature \cite{Yu_2026} \cite{li2021fourierneuraloperatorparametric} \cite{shin2024pseudodifferentialneuraloperatorgeneralized}, such as through the Fourier transform. The primitive versions of these technologies have adapted in breadth and scope upon their subsequent years. Therefore, we are not to be remiss of complex analytic perspectives to artificial intelligence. Newfound results utilizing complex and potentially holomorphic structure are welcoming in contemporary artificial intelligence, thus we find significance in contribution to this area.

\vspace{2mm}

\noindent The property that complex space is isomorphic to twofold Euclidean space is under-influenced, although studied on occasion \cite{barrachina2021equivalence} \cite{barrachina2021complexvaluedvsrealvaluedneural} \cite{Chatterjee_2022} \cite{mönning2018evaluationcomplexvaluedneuralnetworks}, in data science literature. Data structures that hold vanilla information are tangentially related to complex counterparts by simply choosing an even-in-cardinality dimension of the hypothesis class, and reallocating data from a purely Euclidean to complex perspective, notably under an imaginary unit which is widely supported on modern machines such as PyTorch. It is in this notion we find value in treating real data as belonging to a wider hypothesis class, both in the sense of a finite-dimensional parameter space and in pre-activation of where ambient data exists. 

\vspace{2mm}

\noindent It may be expected that the asymptotic ratio of storage capacities is exactly up to a factor of two since the parameter space is affected by a factor of two. The pre-activation and the parameter belong to a finite-dimensional Banach space, thus the parameter's capacity in terms of dimension of this space is exactly affected by two. It is not the case that Gardner volume and their associated quantities are cut in direct respect to this reduction in cardinality. For example, in \cite{xu2025storagecapacityperceptronvariable}, the authors discuss how optimal variable selection can surpass a naïve bound. A bound $\alpha = 2 \rho$ is typical upon selection at random. Our work resides in a geometric projection, thus a deviation from this selection at random as in \cite{xu2025storagecapacityperceptronvariable}. Our work differs from the standard, thus we would also expect our result to cohere to something nonstandard, such as in \cite{xu2025storagecapacityperceptronvariable}. Complex networks exist in literature \cite{Bottcher2024}, although this field is mostly nascent in a spin glass context.

\vspace{2mm}

\noindent Our techniques partially compatible with a complex manifold perspective due to the measure of \ref{eqn:top_form}. In general, we would consider generalized complex ambient spaces which admit complex geometric qualities; however, via the measure of \ref{eqn:top_form},  our considerations are restrictive due to the canonical choice of complex Gaussian. We also consider integration over the compact unitary group, which is a real manifold. Data belonging to a point in a manifold continuum is very well-established under the manifold hypothesis \cite{farghly2025diffusionmodelsmanifoldhypothesis}. Due to these facts, we (rather narrowly due to the aforementioned canonical choice) consider manifolds that admit Hermitian structures with potential adaptability to Kähler manifolds, since we can write \ref{eqn:top_form} with a potential. We remark a complex symplectic manifold is typically Kähler. In fact, there was an open conjecture on existence of complex symplectic manifolds being Kähler \cite{bogomolov1996guan}, although this is not crucial to our work here. 

\vspace{2mm}

Our techniques require a complex Gaussian prior instead of choice of a prototypical complex hypersphere projection. This will allow use of class invariant functions, which in turn give rise to our contributions. Theoretically, this is consistent with traditional Gardner volume literature because of the large sample theory relation
\begin{align}
\frac{1}{N}\|\Theta\|_2^2
\xrightarrow[N\to\infty]{\text{a.s.}} 1, \quad \Theta \sim \mathcal{CN}(0,I_N) .
\end{align}
We provide further discussion in sections \ref{sec:main_result}, \ref{sec:analysis}, \ref{sec:discussion}, and Appendix \ref{app:proofs}.

\vspace{2mm}

\textbf{Our contributions.} We outline our contributions. We find the ratio of storage capacities between complex and real networks. We primarily introduce the following techniques into a Gardner volume analysis:
\begin{enumerate}
\item[$\text{i}.$] the Harish-Chandra-Itzykson-Zuber (HCIZ) formula (not to be mistaken for the HCIZ integral, which has appearances in spin glass literature);
\item[$\text{ii}.$] use of Schur polynomials in the real case and Zonal polynomials in the complex case;
\item[$\text{iii}.$] the Weyl integration formula;
\item[$\text{iv}.$] a Vandermonde matrix;
\item[$\text{v}.$] the Andreief identity.
\end{enumerate}
Our work has general advantages over more classical techniques such as in \cite{gardner1988optimal}. We discuss our contributions.

\vspace{2mm}

\noindent Our strategies maintain high originality due to use of the Harish-Chandra-Itzykson-Zuber (HCIZ) formula \cite{Tao2013HCIZ}, which has not been seen in spin glass literature to our knowledge in the manner we present it. Our overall proof strategy is to compare two scenarios and deduce a final ratio, but this is a end goal and we need a mechanism to compare a final result. The HCIZ formula will provide for us a sharper bound. Thus, we find application of these more obscure strategies mostly vital to our work.

\vspace{2mm}

We highlight reasons to consider the HCIZ formula in spin glass theory. This trick has some appearance in spin glass literature \cite{Kabashima_2008} \cite{Shinzato_2008} \cite{fan2024replicasymmetricfreeenergyising} \cite{maillard2024bayesoptimal}, but generally in the context of Itzykson-Zuber integrals, and not in the manner we present it here. Other references that discuss the HCIZ formula are \cite{husson2023sphericalintegralssublinearrank} \cite{Maillard_2022}.

\vspace{2mm}

\noindent Traditional scenarios of Gardner volume \cite{sorscher2022beyond} \cite{cruciani2022capacityneuralnetworks} evaluation resort to Laplace's method \cite{hughes2025discretelaplaceasymptoticmethod} \cite{_api_ski_2020} to evaluate integration of exponential functions via extremal approximation. It can be noted Laplace's method is largely an asymptotic approximation. In contrast, the HCIZ formula evaluation is exact over the unitary group $U(n)$ for finite $n$. Therefore, the HCIZ formula retains all finite-sized corrections. In particular, our spectral methods preserve information, whereas Laplace's method can destroy it. We will still use Laplace's method, but late in the proof. Applying it late is more robust than applying it early.

\vspace{2mm}

\noindent Laplace's method results in a set of coupled, nonlinear saddle point equations which are nontrivial to solve. We find these strategies complement our use of Schur polynomials \cite{prasad2018introductionschurpolynomials} as well. Under Laplace's method, we must manually attempt to guess the underlying structure via an ansatz. Our methods can overcome that limitation only in early stages. Choosing an ansatz also narrows the scope of the problem, since it is restrictive towards a closed-form solution. Our methods remain as generalized as possible for longer.

\vspace{2mm}

\noindent We elaborate on how our methods use Schur and Zonal polynomials \cite{prasad2018introductionschurpolynomials} \cite{wang2024explicitformulazonalpolynomials}. Traditionally, the evaluation of the limit reduces down to a saddle point problem \cite{zavatoneveth2025expository} \cite{dauphin2014identifyingattackingsaddlepoint}. Taking the limit $n \rightarrow 0$ is nontrivial especially with integration over Hermitian matrices. There is not much room for analytic theory in reducing $n \in \mathbb{Z}$ to $n \rightarrow 0$ over $n \times n$ matrices. With a Schur polynomial decomposition, we can express the limit as a combinatorial weight problem. The closed form limit evaluation will be in terms of Schur weights. Our means of comparison will be to compare final results in terms of these Schur weights, thus the overall limit evaluation strategy is more feasible under our methods.

\vspace{2mm}

\noindent \textbf{Additional discussion.} We discuss how deferring Laplace's method and extremal techniques to late in the proof affects robustness and overall accuracy. Suppose there exists a margin of error as some percentage at an iterate in the proof. Subsequent calculations upon this error can amplify to various scales the margin of error. Moreover, this margin of error can accumulate in further error upon even more approximation (i.e. a saddle point estimation). If this margin is deferred to the end without subsequent calculations, these effects may not be amplified by subsequent calculations. Thus, we explain an effect of error propagation. We support these arguments with Figure \ref{fig:ansatz}.

\vspace{2mm}

\noindent The HCIZ formula is unconventional in literature primarily because it is not widely applicable. Our use of the HCIZ formula requires integration over the unitary compact manifold, which is generated for us from a more conventional domain via change of variables of the Weyl integration formula. Thus, the HCIZ formula is applicable for our contexts.

\vspace{2mm}

\textbf{Related work.} \cite{Gardner_1987}  \cite{gardner1988optimal} \cite{Gardner_1988} are the most clasical references for Gardner volume calculations, relying on saddle approximations, extremal techniques, and Laplace's method. These methods concern the "version space," or the fractional volume of all possible synaptic weights (in the neural sense) in a network that can successfully classify, for example by examining
\begin{align}
V = \frac{1}{V_0} \int_{\mathbb{R}^N} d^N w \prod_{\mu=1}^P \Theta\left( \frac{1}{\sqrt{N}} y^\mu \langle w \cdot \xi^\mu \rangle - \kappa \right) .
\end{align}
Here, $V_0$ is unconstrained volume, $\xi$ is input, $y$ is a label, and $\kappa \in \mathbb{R}$ is a margin. Traditionally, $\Theta$ is the Heaviside step function, although our notation will slightly differ. Gardner introduced the replica trick
\begin{align}
\langle \log V \rangle = \lim_{n \to 0} \frac{\langle V^n \rangle - 1}{n} .
\end{align}
as well as the replica symmetric ansatz. We will discuss these techniques a bit more in throughout this work. 

\vspace{2mm}

The reference with greatest similarity to us to our knowledge is \cite{maillard2024bayesoptimal}, which maintains a similar process of using HCIZ integrals via orthogonally invariant class functions. Their usage is not dependent on Schur and Zonal weights, but our decomposition requires such weights along with the Weyl integration formula. Therefore, while an orthogonally invariant class function process is not new, our exact scenario is. We also remark \cite{maillard2024bayesoptimal} also requires use of the Haar measure. Moreover, \cite{Maillard_2022} with author overlap mentions HCIZ matrix integrals, although this work is more statistical analysis than storage capacity analysis.

\vspace{2mm}

References that discuss use of Schur polynomials in random matrix theory and spin glass are \cite{Matsumoto_2007} \cite{MATSUMOTO_2013}. \cite{Duenez2004} is a more pure work for geometry and measure in random matrix theory, and uses polynomial decompositions but not Schur or Zonal polynomials. \cite{Coulembier_2015} is a closely related work that not only discusses Schur polynomials, but also discusses HCIZ integrals. This work is in random matrix theory and largely discusses rotation invariance under Stiefel manifolds. \cite{AHL_2023__6__975_0} also discusses Stiefel manifolds in random matrix theory and specifically deals with Zonal polynomials.

\vspace{2mm}

\cite{ZHENXIANGCHEN1996157} discusses storage capacities for complex phasor neural networks, although their work is reliant on signal-to-noise theory rather than Gardner volume calculation.

\vspace{2mm}

Non-spherical geometries in generalized Gardner volume calculation exist in literature. \cite{L_F_abbott_1989} consider alternative measures, which allow additional properties away from saturation. \cite{M_Mezard_1989} computes a Gardner quantity using a cavity method and altenerative measures. \cite{Gy_rgyi_2001} discusses replica symmetric breaking and the McCulloch-Pitts neuron, dealing with alternative measures in the process.

\section{Notations and conventions}
\label{sec:notations}

We will use notation
\begin{align}
\int_{\mathcal{H}} dQ \int_{i \mathcal{H}} d \widehat{Q} f(Q,\widehat{Q}) \propto \int_{\mathbb{R}^{n^2}} dx \int_{\mathbb{R}^{n^2}} dy f(x, iy) ,
\end{align}
where $Q \in \mathcal{H}_n$ is in the space of all $n\times n$ Hermitian matrices (a finite-dimensional real manifold isomorphic to $\mathbb{R}^{n^2}$), and $\widehat{Q} \in i \mathcal{H} = \{ i X : X \in \mathcal{H}_n \}$. We use notation
\begin{align}
\int_{\mathbb{C}^N} \exp \left( - \Theta^{\dagger} \Theta \right) \frac{\omega^N}{N!} \quad \text{and} \quad \int_{(\mathbb{C}^N)^n} \left( \bigwedge_{a=1}^n \exp \left( - \Theta_a^{\dagger} \Theta_a \right) \frac{\omega_a^N}{N!} \right)
\end{align}
to denote the integration with respect to the complex Gaussian measure along the $2N$-form on the complex manifold, and the latter to iterate the integral $n$ times. We omit constants in the measure, which are canceled in the final result. 

\vspace{2mm}

We will use terminology \textit{extensive} meaning growing in proportion to $N$, and \textit{macroscopic} to mean a contribution on the scale of the whole system, also scaling in $N$, generally meant to be a more broad term than extensive.

\section{Background}
\label{sec:background}

Our work examines the Gardner volume, a proxy representation for neural network expressivity. Expressivity here means the breadth or capacity a neural network architecture can represent, i.e. neural networks with greater capacity and better theoretical estimators in an empirical, real-world setting. The Gardner volume is typically defined as the quantity \cite{urushibata2024storagecapacityevaluationquantum}
\begin{align}
V = \int \left( \prod_{i=1}^N d\Theta_i \right) \delta\left(\sum_{i=1}^N \Theta_i^2 - N\right) \prod_{\mu=1}^P H\left( \frac{1}{\sqrt{N}} \sum_{i=1}^N \Theta_i \xi_i^\mu - \kappa \right) 
\end{align}
for a single layer continuous perceptron, where there are $N$ weights $\Theta$, input data $\xi$, and a required classification margin $\kappa$. $H$ is a piecewise Heaviside function \begin{align}
H(\nu) = \begin{cases} 1 & \text{if } \nu \ge 0 \\ 0 & \text{if } \nu < 0 . \end{cases} 
\end{align}
We remark the above is often denoted $\Theta$ but we will reserve that our parameter. This integral explores the $N$-dimensional weight space. $V$ is effectively a geometric measure. The Gardner volume $V$ is whatever surface area of the soft or hard shell hypersphere remains after $P$ intersecting half-spaces have chopped away the invalid regions. If $V$ is large, the region of valid weights is vast, i.e. expressive. If $V$ is small, or shrinks as more data is added, more cuts are made, and learning is hard because a microscopic target is developed.

\vspace{2mm}

\noindent We examine critical capacities $\alpha_{\text{critical}}$. As we increase $P$, the phase space of valid weights shrinks because the weights must satisfy more constraints. The critical capacity $\alpha_{\text{critical}}$ is the threshold in which the Gardner volume collapses to zero. When $\alpha < \alpha_{\text{critical}}$, a solution exists and the network can memorize the dataset. When $\alpha > \alpha_{\text{critical}}$, no set of weights exists that can satisfy all patterns, and training loss is positive in an ideal scenario. It is a famous result that with $\kappa=0$ that $\alpha_{\text{critical}} = 2$ \cite{Gardner_1988}. Typical methods for computing $\alpha_{\text{critical}}$ involve the replica trick \cite{gardner1988optimal} and noting $\langle \log V \rangle = \lim_{n \to 0} \frac{\langle V^n \rangle - 1}{n}$. Because we operate in the thermodynamic limit and so $N \rightarrow \infty$, Laplace's method is used.

\section{Main result}
\label{sec:main_result}

Let us provide the main result, then we supplement the main result and provide a discussion.

\vspace{2mm}

\noindent \textit{\textbf{Theorem 1.} Suppose sufficient exchange of limit order is allowed (\cite{Barra_2012} \cite{Coolen_2017}). Let $\Xi \subseteq \mathbb{C}^n$ be the parameter space of a complex-valued neural network with weights $\Theta \in \Xi$, equipped with a natural Hermitian structure corresponding to that constrained to the complex unit Gaussian. Assume $\theta_{\mu}^{\lambda}$ as in Appendices \ref{app:complex}, \ref{app:proofs} takes the desired form of \ref{eqn:theta_form}. Let $\alpha_r$ denote the storage capacity of the complex hypothesis class subject to this real pre-activation constraint, and let $\alpha_c$ denote the storage capacity of the unconstrained complex hypothesis class. Then with high probability
\begin{align}
\frac{\alpha_r}{\alpha_c} = \Vast( \lim_{q \rightarrow 1} \frac{\frac{q}{1-q} - \mathcal{G}_r'(q)}{\displaystyle \frac{\partial}{\partial q} \int \mathcal{D}t \log \Phi\left( \frac{\sqrt{q} t - \kappa}{\sqrt{1-q}} \right)} \Vast) \Bigg(  \lim_{\rho \rightarrow \rho_c} \frac{\frac{\partial}{\partial \rho} \mathcal{G}_c(\rho,\kappa)}{1  - 2 \log 2 + \log ( 1 + \rho )  - \mathcal{B}'(\rho)}  \Bigg)  := \Gamma ,
\end{align}
$0 \leq \Gamma \leq 1$, in the asymptotic limit, where $\kappa$ is a generalization margin, $\rho \in (0,1]$ is a sparsity constraint parameter, $\mathcal{G}_r, \mathcal{G}_c \neq 0$ are saddle point evaluations, $\mathcal{B}$ is a residual term under a combinatorial asymptotic expansion, and $\Phi$ is the standard normal cumulative distribution function.}

\vspace{2mm}

A major assumption, which is notable since it slightly deviates from literature, we will make is the totality of weights is drawn from
\begin{align}
\Theta \sim \mathcal{C} \mathcal{N}(0,I_N) .
\end{align}
This will allow $G_0$ class invariance, but not total class invariance, which we see in equations \ref{eqn:U_invariance}, \ref{eqn:O_invariance}. Note that this setup preserves similar scaling as the sphere, which is the canonical choice for Gardner volume parameter setup. Geometrically, we are replacing a hard shell around $\| \Theta \| = \sqrt{N}$ with now a "soft shell," because that is where the Gaussian concentrates in norm. In particular, the substitution of the complex Gaussian over the complex hypersphere is geometrically valid because the Gaussian distribution naturally concentrates its mass at this exact boundary in high-dimensional spaces. Without this application, our class invariances fail. The final calculation uses the fact that for the Gaussian prior $\Theta \sim \mathcal{CN}(0, I_N)$, we get
\begin{align}
\label{eqn:high_prob}
q_d = \frac{1}{N} \|\Theta\|_2^2 \xrightarrow{N \to \infty} 1 \quad \text{almost surely} ,
\end{align}
where $q_d$ is the diagonal of the ansatz. The distribution itself is indeed centered at zero, meaning the expected value of the weights is 0; however, the norm of a vector drawn from this distribution concentrates near 1. Moreover, a complex Gaussian is compatible with the canonical flat metric (we refer to Figure \ref{fig:weights_and_metric}). In reference to \ref{eqn:high_prob}, recall the hypersphere condition $\| \Theta \|_2 = \sqrt{N}$, and \ref{eqn:high_prob} is consistent with that by rearranging and taking the square root. We refer to Theorem 2 in \ref{sec:discussion} for theoretical evidence our strategies have commonground with traditional calculation.

\vspace{2mm}

\subsection{Subsidiary results}
\label{sec:subsidiary}

Before we analyze the main result, we provide discussion of subsidiary results to support the proof. We begin with the claim:

\vspace{2mm}

\noindent \textit{\textbf{Claim}. We have
\begin{align}
\label{eqn:claim_statement_main}
1 & = \lim_{\epsilon \to 0^+} \int_{\mathcal{H}_n} dQ \int_{i\mathcal{H}_n} d\widehat{Q}  \exp\Bigg( -N \text{Tr}\Bigg( \widehat{Q} (Q - R(\Theta)) \Bigg) \Bigg) \exp\Bigg(-\epsilon \text{Tr}(Q^2 - \widehat{Q}^2)\Bigg) ,
\end{align} 
where $Q \in \mathcal{H}_n$ is in the space of all $n\times n$ Hermitian matrices, and $\widehat{Q} \in i \mathcal{H}_n = \{ i X : X \in \mathcal{H}_n \}$, and $R(\Theta) \in \mathcal{H}_n, R_{ab}(\Theta) = \frac{1}{N} \Theta_a^{\dagger} \Theta_b$.}

\begin{figure}[!t]
  \centering
  \includegraphics[width=0.6\linewidth]{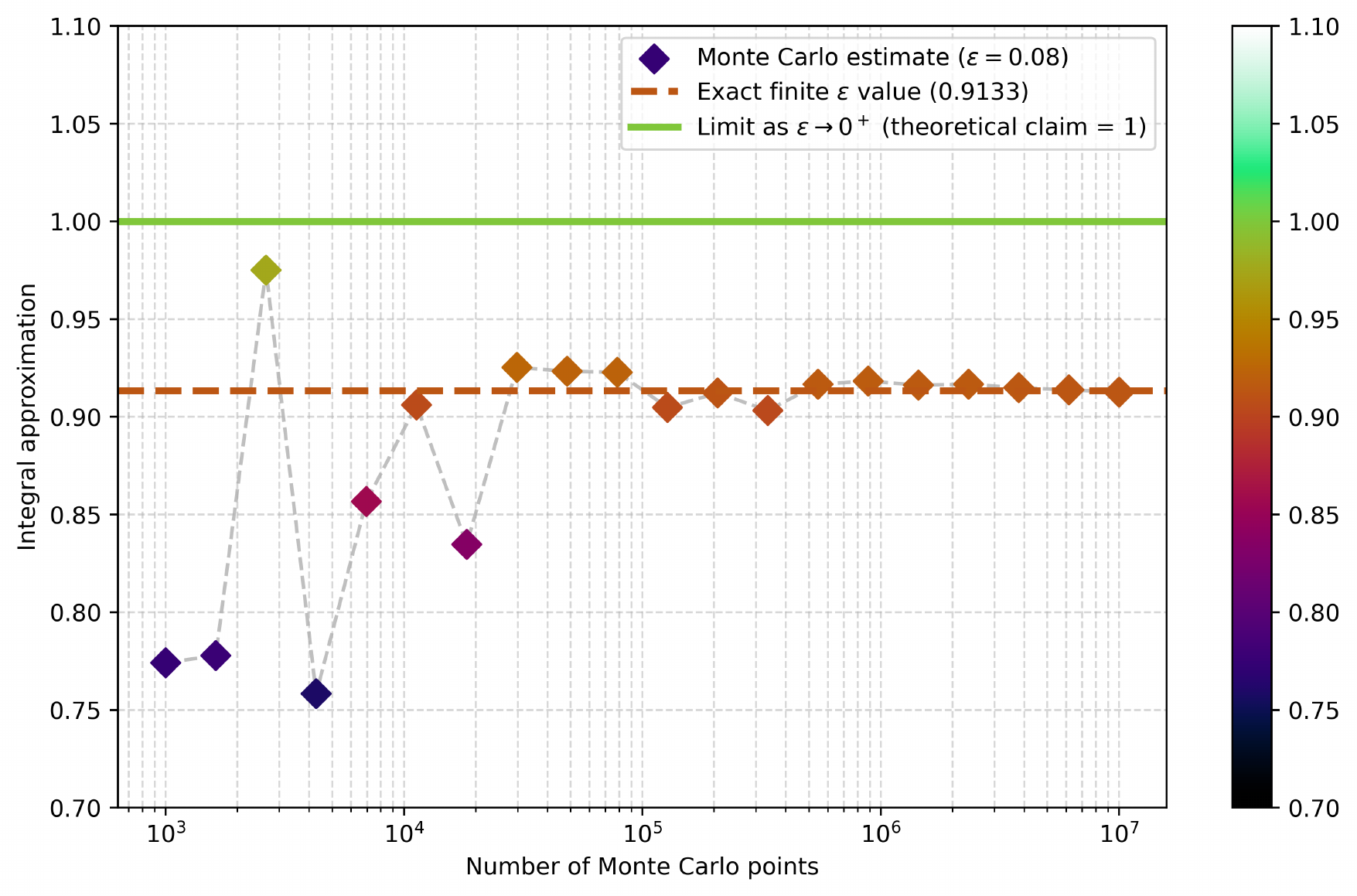}
  \vspace{-2mm}
  \caption{We plot the claim of \ref{sec:subsidiary} in a real scenario via Monte Carlo integration. We reduce dimensionality to $n=1$ for simplicity. We plot asymptotics of $I(\epsilon) = \frac{N}{2\pi} \int_{-\infty}^\infty dQ \int_{-\infty}^\infty dY \exp(-i N Y (Q - R)) \exp(-\epsilon Q^2 - \epsilon Y^2)$. Because the sine term of the complex exponential evaluates to $0$ by symmetry, we only need to evaluate the real part $I(\epsilon) = \frac{N}{2\epsilon} \mathbb{E}_{Q, Y \sim \mathcal{N}(0, \frac{1}{2\epsilon})} \Big[ \cos(N Y (Q - R)) \Big]$. To evaluate the claim, we choose fixed $\epsilon = 0.08$ before the limit threshold rather than taking the limit.}
  \label{fig:claim}
\end{figure}

\vspace{2mm}

The claim is a multidimensional generalization of the Fourier representation of the Dirac delta function. The core of the claim is the identity
\begin{align}
\label{eqn:dirac_fourier_main}
\frac{1}{2\pi} \int_{-\infty}^{\infty} dk e^{ikx} = \delta(x) .
\end{align}
This claim is reminiscent of what appears in random matrix theory. A limit followed by integrating of this form often appears in the context of the Hubbard-Stratonovich transformation. The matrix $Q$ in \ref{eqn:claim_statement_main} acts as the "position" variable, $\widehat{Q}$ acts as the "momentum" or "frequency" variable, and the small parameter $\epsilon$ is a regularizer that makes the highly oscillatory integrals mathematically well-defined before taking the limit $\epsilon \to 0^+$ just after the equality in \ref{eqn:claim_statement_main}. The claim relies on the setup:
\begin{enumerate}
\item[$\text{i}.$] $Q$ is the matrix analog of $x$. $R(\Theta)$ is a constant shift. Therefore, $X = Q - R(\Theta)$ is the shifted position in the trace of \ref{eqn:claim_statement_main}.
\item[$\text{ii}.$] Because $\widehat{Q} \in i\mathcal{H}_n$, it can be written as $iY$ for some Hermitian matrix $Y$. This provides the imaginary unit $i$ needed in the exponent, making $\text{Tr}(\widehat{Q}X)$ analogous to $ikx$ as in \ref{eqn:dirac_fourier_main}.
\item[$\text{iii}.$] Highly-oscillatory integrals $\int dk e^{ikx} $ are not absolutely convergent. In order to evaluate, we multiply by an anologous decaying Gaussian $e^{-\epsilon k^2}$ and take the limit as $\epsilon \to 0$. Because $\widehat{Q}$ is anti-Hermitian, its square $\widehat{Q}^2$ is negative (semi-)definite. Thus, $e^{\epsilon \text{Tr}(\widehat{Q}^2) }$ is a decaying Gaussian regulator.
\end{enumerate}
This claim has connections to a nascent Dirac delta. Let us examine this more closely. The proof relies on an integral of the form
\begin{align}
I_\epsilon(X_{aa}) = \frac{N}{2\pi i} \int_{-i\infty}^{i\infty} d\widehat{Q}_{aa} \exp\Big(-N \widehat{Q}_{aa} X_{aa} + \epsilon \widehat{Q}_{aa}^2\Big)  ,
\end{align}
which is essentially the Fourier transform of a Gaussian. Using the identity $\int dy e^{-ay^2 + by}  = \sqrt{\frac{\pi}{a}} e^{b^2 / 4a}$ with $a = \epsilon$ and $b = -i N X_{aa}$
\begin{align}
\label{eqn:I}
I_\epsilon(X_{aa}) = \frac{N}{2\pi} \sqrt{\frac{\pi}{\epsilon}} \exp\left( \frac{(-i N X_{aa})^2}{4\epsilon} \right)  = \frac{N}{2\sqrt{\pi \epsilon}} \exp\left( - \frac{N^2 X_{aa}^2}{4\epsilon} \right) .
\end{align}
Via definition of the nascent Dirac delta, it is a parameterized family of locally integrable functions $\eta_\epsilon(x)$ such that, for any continuous compactly supported test function $\phi$, the sequence converges in the weak-* topology of distributions
\begin{align}
\label{eqn:nascent_eta}
\lim_{\epsilon \to 0^+} \int dx \eta_\epsilon(x) \phi(x)  = \phi(0) .
\end{align}
If we define a variance parameter $\sigma^2 = \frac{2\epsilon}{N^2}$, our evaluated integral $I_\epsilon(X_{aa})$ of \ref{eqn:I} rewrites as the canonical normal density
\begin{align}
I_\epsilon(X_{aa}) = \frac{1}{\sqrt{2\pi \sigma^2}} \exp\left( - \frac{X_{aa}^2}{2\sigma^2} \right) .
\end{align}
As $\epsilon \to 0^+$, the variance $\sigma^2 \to 0$. The Gaussian becomes infinitely tall and narrow at $X_{aa} = 0$, while its integral over $\mathbb{R}$ remains $1$ for all $\epsilon > 0$. By doing this across all $n^2$ independent real coordinates (diagonal, real off-diagonal, and imaginary off-diagonal), the full inner integral evaluates to a product of $n^2$ independent Gaussian nascent deltas using notation of \ref{eqn:nascent_eta}
\begin{align}
\eta_\epsilon(Q - R(\Theta)) \propto \exp\left( - \frac{N^2}{4\epsilon} \text{Tr}\Big((Q - R(\Theta))^2\Big) \right) .
\end{align}

\vspace{2mm}

Now, we provide the following three lemmas, which will help us in the proof of the main result.

\vspace{2mm}

\textit{\textbf{Lemma 1.} Let $\mathcal{S}_{\text{branch}}$ be as in \ref{app:complex}. It follows that
\begin{align}
\mathcal{S}_{\text{branch}} = -l^* \log(l^*) + l^* \log(N) + \mathcal{B}(l^*) .
\end{align}}

\vspace{2mm}

\textit{\textbf{Lemma 2.} Define $\mathcal{S}_{\text{patterns}}$ as in \ref{app:complex}. It follows that
\begin{align}
\mathcal{S}_{\text{patterns}}(\alpha_c) = -\rho N \log N + N v(Q^*) + \alpha_c N \mathcal{G}_c +  o(N) .    
\end{align}}

\vspace{2mm}

\textit{\textbf{Lemma 3.} Define $\mathcal{S}_{\text{prior}}$ as in \ref{app:complex}. It follows that
\begin{align}
\mathcal{S}_{\text{prior}}(\alpha_c) = (1+\rho)\log(1+\rho) - \rho\log\rho .    
\end{align}}

\vspace{2mm}

The supporting lemmas are relevant to a decomposition as in Laplace's method, in the proof the line being \ref{eqn:laplace_complex}
\begin{align}
\sum_{\lambda} \sum_{\mu} \exp \Big\{ S_{\text{unconstrained}}(\lambda, \mu, n) \Big\} \xrightarrow[\displaystyle \propto]{N \to \infty} \exp \Big\{ \text{extr} \underbrace{  S_{\text{unconstrained}}(\lambda^*, \mu^*, n)  }_{\displaystyle = \sum_i \mathcal{S}_i} \Big\} ,
\end{align}
and $\mathcal{S}_{\text{branch}}, \mathcal{S}_{\text{patterns}}, \mathcal{S}_{\text{prior}}$ are terms in the decomposition/sum $\sum_i \mathcal{S}_i$. The proofs of the lemmas primarily stem from
\begin{enumerate}
\item[$\text{i}.$] Lemma 1. It follows from a change-of-basis
\begin{align}
\label{eqn:theta_form}
\theta_{\mu^*}^{\lambda^*}(n) \approx  \left( \frac{N^{l^*}}{l^*!} \right)^n   e^{n \widetilde{\mathcal{B}}(l^*)}   ,
\end{align}
which first appears in \ref{eqn:change_basis} via Zonal polynomial term interaction.
\item[$\text{ii}.$] Lemma 2. It begins with a Zonal polynomial representation via Gibbs weights \begin{align}
c_{\lambda^*}(P) = \mathbb{E}_{\xi} \left[ \int d\mu_{\lambda^*}(\Theta^1, \dots, \Theta^n) \prod_{\mu=1}^P \prod_{a=1}^n \mathbb{X}(\Theta^a, \xi^\mu) \right]
\end{align} 
and mostly manipulating this equation, notably in the thermodynamic limit.
\item[$\text{iii}.$] Lemma 3. The prior term as in \ref{eqn:laplace_complex} can be reduced to in the limit
\begin{align}
\frac{1}{N} S_{\text{prior}} \longrightarrow (1+\rho)\log(1+\rho) - \rho\log\rho ,
\end{align}
which closely align with strategies in section \ref{sec:analysis}.
\end{enumerate}

\subsection{Analysis and closed forms of main result}
\label{sec:analysis}

The ostensible pitfall of the main result is that $\mathcal{G}_r, \mathcal{G}_c, \mathcal{B}(\rho)$ have nontrivial closed forms. We attempt to simplify this and develop a more closed-form solution.

\vspace{2mm}

\begin{figure}[!t]
  \centering
  \includegraphics[width=0.85\linewidth]{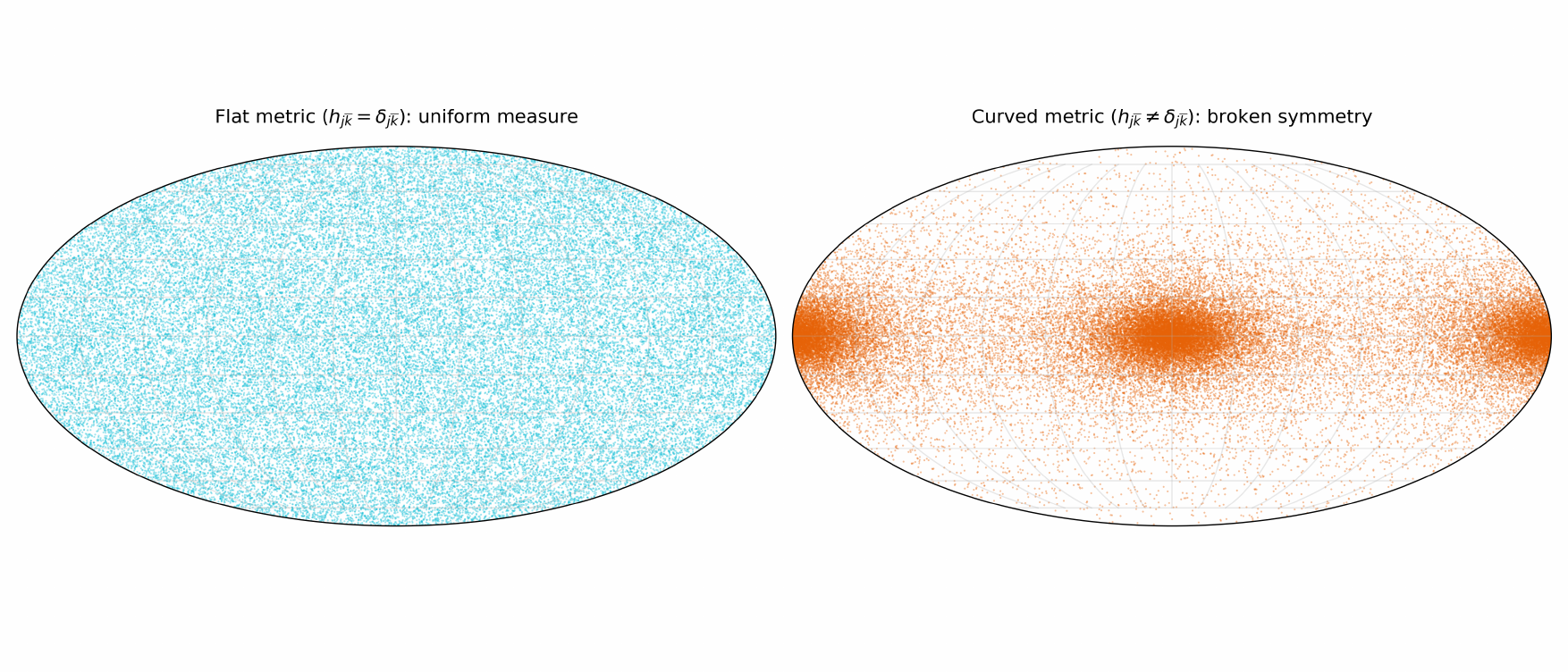}
  \vspace{-2mm}
  \caption{We illustrate weights across a hypersphere projection according a random initialization of a network. The left illustrates a uniform, unbiased initialization, which correspond to a flat metric due to the uniformity. In particular, a weight vector $\Theta = (\theta_1, \dots, \theta_N)$ is equally likely to point in any direction. This argument has greater correspondence to a flat metric pairing with a hyperspherical geometry, and we remark our choice of complex Gaussian is canonically associated with the flat metric. In the right plot, we sample according to an anisotropic prior, thus a curved metric, and the initialization is nonuniform.}
  \label{fig:weights_and_metric}
\end{figure}

The aforementioned terms rely on choice of ambient metric, or volume top form that we develop in Appendix \ref{app:real_phase} via \ref{eqn:top_form}, which is
\begin{align}
\label{eqn:top_form}
d\mu(\Theta) = \frac{1}{Z} e^{-K(\Theta, \overline{\Theta})} \det(h_{j\overline{k}}) \left(\frac{i}{2}\right)^N \bigwedge_{\ell=1}^N d\theta_{\ell} \wedge d\overline{\theta}_{\ell}  .
\end{align}
$Z$ is a normalization constant. This choice largely lacks freedom because of the complex Gaussian consideration that follows this construction in Appendix \ref{app:real_phase}. From a physical perspective, setting our ambient structure to the canonical flat metric $h_{j\overline{k}} = \delta_{j\overline{k}}$ is compatible with complex Gaussian weight initialization, whose density depends on the radial quantity $\Theta^{\dagger} \Theta$. Moreover, we get
\begin{align}
\EX \| \Theta \|^2 = N ,
\end{align}
which implies $\| \Theta \| \approx \sqrt{N}$ up to expected value. When we constrain these weights to the soft shell complex hypersphere, an isotropic prior creates a uniform probability measure across the manifold. In our case, the weights are not constrained exactly; however, the Gaussian model retains scale of the hypersphere, which is exactly compatible with the flat metric. We support general compatibility Figure \ref{fig:weights_and_metric}.

\vspace{2mm}

The term $\mathcal{G}_r(q)$ originates from, by definition,
\begin{align}
\mathcal{G}_r(q) := \lim_{N \to \infty} \frac{1}{N} \left. \frac{\partial \log c_{\lambda^*}(N)}{\partial n} \right|_{n=0} .
\end{align}
Here, $c_{\lambda^*}(N)$ is defined such that $\exp(c_{N,P} G(A)) = \sum_{\lambda} c_\lambda(P) s_\lambda(A)$ is the Schur polynomial decomposition. In Appendix \ref{app:real_phase}, we have
\begin{align}
\label{eqn:G0}
\exp\Big(N G_0(\widehat{Q})\Big) := \int_{(\mathbb{C}^N)^n} \left( \bigwedge_{a=1}^n \exp \left( - \Theta_a^{\dagger} \Theta_a \right) \frac{\omega_a^N}{N!} \right) \exp \Bigg( N \text{Tr} \Big( \widehat{Q} R(\Theta) \Big)\Bigg) ,
\end{align}
which can be observed by following through \ref{eqn:G0} and \ref{eqn:real_Vn_decomposed}, with corresponding power series
\begin{align}
1 + N \text{Tr}(\widehat{Q} R(\Theta)) + \frac{N^2}{2!} \big(\text{Tr}(\widehat{Q} R(\Theta))\big)^2 + \dots 
\end{align}
can be integrated with respect to the Gaussian measure, giving (maintaining all order, since the exponential captures the entire power series representation)
\begin{align}
\int_{\mathbb{C}^{n \times N}} d\Theta d\overline{\Theta} \exp \left(-\text{Tr}(\Theta^\dagger \Theta) \right) \exp \left(\text{Tr} (\widehat{Q} \Theta^\dagger \Theta) \right)  \propto \det(I - \widehat{Q})^{-N} ,
\end{align}
which follows since 
\begin{align}
\int_{\mathbb{C}^{N \times K}} d\Theta d\overline{\Theta} \exp\left( -\sum_{j=1}^N \widetilde{\theta}_j^\dagger (I_K - \widehat{Q}) \widetilde{\theta}_j \right) = \prod_{j=1}^N \left( \int_{\mathbb{C}^K} d\widetilde{\theta}_j d\overline{\widetilde{\theta}}_j \exp\left( -\widetilde{\theta}_j^\dagger (I_K - \widehat{Q}) \widetilde{\theta}_j \right)  \right) , 
\end{align}
and by evaluating via a Gaussian integral. Recall $\int_{\mathbb{C}^K} \exp( -z^\dagger A z ) dz d\overline{z} = \frac{\pi^K}{\det(A)}$, therefore the formula is reminiscent of this. By the Cauchy-Littlewood identity (see \cite{macdonald1995symmetric} and \cite{WHEELER2016543} for discussion), this expands into Schur polynomials as
\begin{align}
\label{eqn:cauchy_littlewood}
\det(I - \widehat{Q})^{-N} = \sum_{\lambda} s_\lambda(I_N) s_\lambda(\widehat{Q}) ,
\end{align}
where $s_\lambda(I_N)$ is the dimension of the irreducible representation of the general linear group $\text{GL}(N, \mathbb{C})$ associated with the partition $\lambda$. To find $c_\lambda(N)$, we "coefficient match" and we deduce
\begin{align}
\label{eqn:coefficient_matching}
c_\lambda(N) = s_\lambda(I_N)  .
\end{align} 
We refer to \cite{macdonald1995symmetric} for relevant literature. Now, via the Weyl dimension formula \cite{fulton1991representation}, $s_{\lambda}(I_N)$ using \ref{eqn:cauchy_littlewood} can be written
\begin{align}
s_\lambda(I_N) =  \prod_{1 \leq i < j \leq N} \frac{\lambda_i - \lambda_j + j - i}{j - i} .
\end{align}

\vspace{2mm}

Now, we examine $\mathcal{G}_c$. By definition, 
\begin{align}
\label{eqn:G_c_definition}
\mathcal{G}_c(Q^*) := \lim_{n \to 0} \frac{1}{n} \log \mathbb{E}_{\xi} \left[ \prod_{a=1}^n \mathbb{X}(\Theta^a, \xi) \right]_{ \frac{1}{N}\Theta^\dagger \Theta = Q^*} = \lim_{n \to 0} \frac{1}{n} \log \mathbb{E}_{\xi} \left[ \prod_{a=1}^n H(\text{Re}(\Psi_a) - \kappa) \right]_{ \frac{1}{N}\Theta^\dagger \Theta = Q^*}.
\end{align}
where $\mathbb{X}$ is a Gibbs weight corresponding to the Heaviside function with pre-activations $\Psi_a = \frac{1}{\sqrt{N}} (\Theta^a)^\dagger \xi$. For suitable covariance matrix $\Sigma_{\rho}$, and setting $u_a = \text{Re}(\Psi_a)$,
\begin{align}
\label{eqn:G_c_integral}
\mathbb{E}_{\xi} \left[ \prod_{a=1}^n H(u_a - \kappa) \right]_{Q^*} = \int_{\mathbb{R}^n}  du \mathcal{N}(u; 0, \Sigma_\rho) \prod_{a=1}^n H(u_a - \kappa)   .
\end{align}
To evaluate this integral, we apply a Hubbard-Stratonovich-type argument. First apply the change of variables $u_a = \frac{1}{\sqrt{2}} [ \sqrt{\rho} t + \sqrt{1-\rho} z_a ]$. \ref{eqn:G_c_integral} can be written using $Q_{aa} \rightarrow 1$
\begin{align}
\mathbb{E}_{\xi} \left[ \prod_{a=1}^n H(u_a - \kappa) \right]_{Q^*} = \int_{-\infty}^\infty \mathcal{D}t \prod_{a=1}^n \left( \int_{-\infty}^\infty \mathcal{D}z_a H\left(\frac{1}{\sqrt{2}} [ \sqrt{\rho} t + \sqrt{1-\rho} z_a ]- \kappa\right) \right) ,
\end{align}
where $\mathcal{D}z = \frac{1}{\sqrt{2\pi}} \exp(-z^2/2) dz$. The inner integral over $z_a$ along with use of the Heaviside function dictates that $\sqrt{1-\rho} z_a \ge \sqrt{2} \kappa- \sqrt{\rho} t \implies z_a \ge \frac{\sqrt{2} \kappa- \sqrt{\rho} t}{\sqrt{1-\rho}}$. We have under the nondegenerate scenario
\begin{align}
\int_{\frac{\sqrt{2}  \kappa- \sqrt{\rho}t}{\sqrt{1-\rho}}}^\infty \mathcal{D}z_a = \Phi\left(\frac{\sqrt{\rho} t - \sqrt{2} \kappa}{\sqrt{1-\rho}}\right) .
\end{align}
Substituting into \ref{eqn:G_c_definition}, we get
\begin{align}
\mathcal{G}_c(Q^*) & = \lim_{n \to 0} \frac{1}{n} \log \left( \int_{-\infty}^\infty \mathcal{D}t \left[ \Phi\left(\frac{\sqrt{\rho} t - \sqrt{2} \kappa}{\sqrt{1-\rho}}\right) \right]^n \right) 
\\
& \stackrel{(1)}{=} \lim_{n \to 0} \frac{1}{n} \log \left( \int_{-\infty}^\infty \mathcal{D}t \left( 1 + n \log \Phi\left(\frac{\sqrt{\rho} t - \sqrt{2} \kappa}{\sqrt{1-\rho}}\right) \right) \right)
\\
& = \lim_{n \to 0} \frac{1}{n} \log \left( 1 + n \int_{-\infty}^\infty \mathcal{D}t \log \Phi\left(\frac{\sqrt{\rho} t - \sqrt{2} \kappa}{\sqrt{1-\rho}}\right) \right)
\\
& \stackrel{(3)}{=} \int_{-\infty}^\infty \mathcal{D}t \log \Phi\left(\frac{\sqrt{\rho} t - \sqrt{2} \kappa}{\sqrt{1-\rho}}\right) .
\end{align}
(1) utilizes $x^n = \exp(n \log x) = 1 + n \log x + \mathcal{O}(n^2)$ and (2) utilizes $\log(1+x) \approx x$ for small $x$.

\vspace{2mm}

We finally examine the term $\mathcal{B}(\rho)$. Recall from Lemma 1 in \ref{sec:subsidiary}, \ref{app:proofs} that the branching contribution takes the form
\begin{align}
\label{eqn:sbranch_analysis}
\mathcal{S}_{\mathrm{branch}} = -l^*\log l^* + l^*\log N + \mathcal{B}(l^*), \qquad \mathcal{B}(l^*) = l^*+\widetilde{\mathcal{B}}(l^*),
\end{align}
where $\widetilde{\mathcal{B}}(l^*)$ is the residual contribution appearing in the asymptotic branching-coefficient assumption of \ref{eqn:theta_form}. In the replica-symmetric thermodynamic scaling, we set $l^*=\rho N$. Hence we rewrite \ref{eqn:sbranch_analysis}
\begin{align} \mathcal{S}_{\mathrm{branch}} &= -\rho N\log(\rho N) + \rho N\log N + \mathcal{B}(\rho N) = -\rho N\log\rho + \mathcal{B}(\rho N). \end{align}
Since the $f$ term of \ref{app:complex} contains $N^{-1}\mathcal{S}_{\mathrm{branch}}$, extensivity requires that $\mathcal{B}(\rho N)$ grow at most linearly in $N$. We therefore define the thermodynamic branching contribution by
\begin{align}
\label{eqn:Brho_def}
\mathcal{B}(\rho) := \lim_{N\to\infty} \frac{1}{N}\mathcal{B}(\rho N).
\end{align} 
Using $\mathcal{B}(l^*)=l^*+\widetilde{\mathcal{B}}(l^*)$, we obtain from \ref{eqn:Brho_def}
\begin{align} \mathcal{B}(\rho) &= \lim_{N\to\infty} \frac{1}{N} \left[ \rho N+\widetilde{\mathcal{B}}(\rho N) \right] = \rho + \lim_{N\to\infty} \frac{1}{N}\widetilde{\mathcal{B}}(\rho N). \end{align}
The first term is the extensive bulk contribution from Stirling's formula in Lemma 1 in Appendix \ref{app:proofs}, while $\widetilde{\mathcal{B}}$ contains the remaining volume correction. This decomposition is exactly how $\mathcal{B}$ arises in the proof of Lemma 1. To obtain a closed form, we further assume that the residual contribution is subextensive in the thermodynamic limit,
\begin{align}
\widetilde{\mathcal{B}}(\rho N) = o(N), \qquad \text{or equivalently,} \qquad \lim_{N\to\infty} \frac{1}{N}\widetilde{\mathcal{B}}(\rho N) = 0.
\end{align}
Thus only the macroscopic bulk term survives in $f$, giving
\begin{align}
\mathcal{B}(\rho) = \rho.
\end{align}

\subsection{Additional discussion of main result}
\label{sec:discussion}

\begin{figure}[!t]
  \centering
  \includegraphics[width=0.95\linewidth]{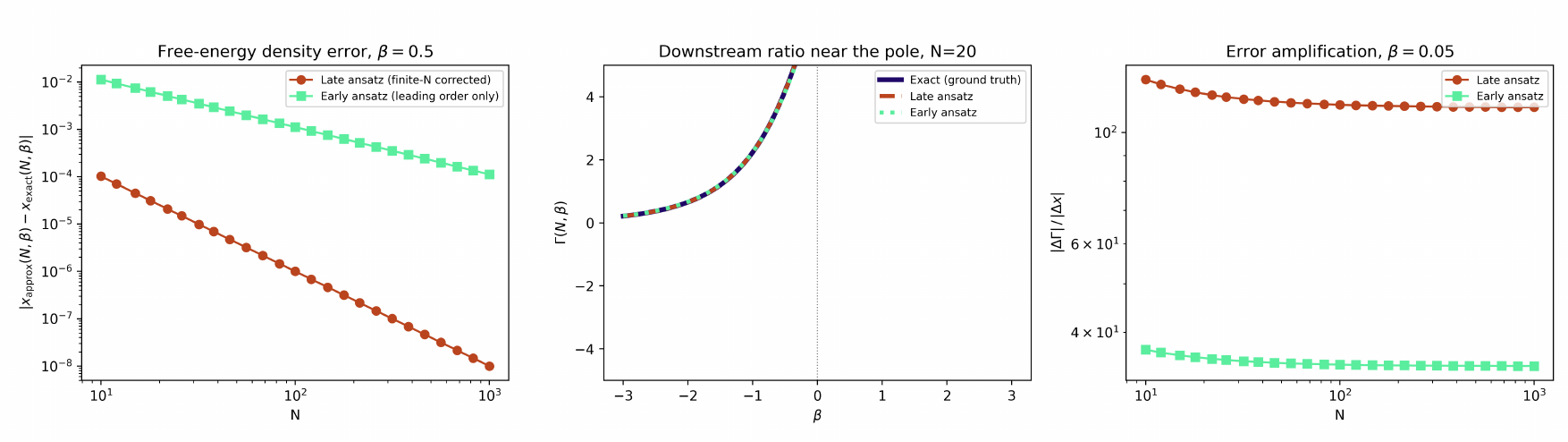}
  \vspace{-2mm}
  \caption{We illustrate that deferring application of the ansatz until late is more robust than applying it early. This figure is partially toy, and is not done with real networks, but rather done with a closed-form numerical evaluations of a partition function $Z = \sum_{l=0}^N \binom{N}{l} \exp(\beta l - \Gamma l^2 / N)$, a 1D scalar equivalent of the high-dimensional Gardner volume calculations $\langle V^n \rangle$. $Z$ is not directly used in the proofs in Appendices \ref{app:real_phase}, \ref{app:complex}; instead, $Z$ requires evaluation based on classical Gardner techniques that variously appear in \ref{app:real_phase}, \ref{app:complex}. The left panel shows absolute error $x$, which has rough correspondence to $f$ as in this work, but $x$ does not rely on the replica trick. In particular, $x_{\text{exact}} = \frac{1}{N} \log Z$. The plot indicates an application of Laplace's method, and the late application carries one order further, and is indicative that the HCIZ formula retains finite-sized corrections. Lower is better. The middle panel shows $\Gamma$ corresponding to a toy version of Theorem 1 is roughly equal for all three scenarios: exact, early-applied, late-applied. This panel sets us up for the right panel. The right panel shows error amplification $|\Delta \Gamma| / |\Delta x|$ at fixed $\beta$, which shows subsequent calculations upon this error can amplify to various scales the margin of error. $\beta$ in this plot is closest to $\widehat{Q}$ as in \ref{app:real_phase}, \ref{app:complex}. Lower is better.}
  \label{fig:ansatz}
\end{figure}

Before we provide a discussion, we supplement the main theorem. We further show our theory is consistent with more traditional Gardner volume methods.

\vspace{2mm}

\textit{\textbf{Theorem 2.} Let $\mathbb{S}_N = \{\Theta \in \mathbb{C}^N \mid \|\Theta\|_2^2 = N\}$ and define the uniform spherical measure $d\mu_{S}(\Theta) \propto \delta(\|\Theta\|_2^2 - N) d\Theta$. Let $d\mu_{G}(\Theta) = \pi^{-N} \exp(-\Theta^\dagger \Theta) d\Theta$ be the complex Gaussian measure under the flat metric $h_{j\overline{k}} = \delta_{j\overline{k}}$. Suppose for a single pattern $z^\mu \sim \mathcal{CN}(0, I_N)$ , and let $\mathbb{X}(\Theta, z^\mu) = H(\text{Re}(\Psi^\mu) - \kappa)$. Assume that $Q_{aa}^* = 1 + o(1)$ in the $d\mu_G$ case, which is necessary because otherwise the Gaussian model would concentrate at a different radius than the spherical model. In the thermodynamic limit $N, P \to \infty$ with $P/N = \alpha$, the quenched free energy densities evaluate to the same
\begin{align} \lim_{N \to \infty} f_{S} = \lim_{N \to \infty} f_{G} . \end{align}}

\vspace{2mm}

\textit{Remark.} Theorem 2 is our bridge that our assumptions belonging the complex Gaussian weights have to varying depth are consistent with more traditional Gardner volume calculations. In particular, our main result in section \ref{sec:main_result} employs use of the replica ansatz in Appendix \ref{app:real_phase}, and our rigorous proofs rely somewhat on shorthand facts relating to $\frac{1}{N} \| \Theta \|_2^2 \rightarrow 1$. Therefore, this theorem formalizes that a bit more.

\vspace{2mm}

Now we discuss the main result further. We omit further outlining contributions, since we did this in \ref{sec:introduction}, therefore we refer to \ref{sec:introduction} for the primary effects of our main proofs and why our contributions are nontrivial.

\vspace{2mm}

The overarching key to why this proof follows is the method in which we begin the proof via the definition of Gardner volume. In the real case and complex cases respectively, we approach the problem with the quenched average
\begin{align}
\begin{cases}
& \langle V_{\text{real}}^n \rangle = \int_{(\mathbb{C}^N)^n} \left( \bigwedge_{a=1}^n \exp \left( - \Theta_a^{\dagger} \Theta_a \right) \frac{\omega_a^N}{N!} \right) \left[ \left\langle \prod_{a=1}^n H\Big(\text{Re}(\Psi_a) - \kappa\Big) \delta\Big(\text{Im}(\Psi_a^\mu)\Big) \right\rangle_Z \right]^P 
\\
& \langle V_{\text{complex}}^n \rangle = \int_{(\mathbb{C}^N)^n} \left( \bigwedge_{a=1}^n \exp \left( - \Theta_a^{\dagger} \Theta_a \right) \frac{\omega_a^N}{N!} \right) \left[ \left\langle \prod_{a=1}^n H\Big(\text{Re}(\Psi_a^\mu) - \kappa\Big) \right\rangle_Z \right]^P 
.
\end{cases}
\end{align}

It is crucial that the above considers complex-valued data, otherwise our methods fail. The Dirac delta fundamentally restricts the version space to \begin{align}
V_r = \left\{ \Theta \mid \text{Re}(\Psi^\mu) \ge \kappa \text{ and } \text{Im}(\Psi^\mu) = 0 \text{ for all } \mu \in [ P ] \right\} ,
\end{align}
and in the complex case
\begin{align}
V_c = \left\{ \Theta  \mid \text{Re}(\Psi^\mu) \ge \kappa \text{ for all } \mu \in [P] \right\}  .
\end{align}
Clearly, the real-constrained version space is a subset of the unconstrained version space $
V_r \subseteq V_c $,
and it immediately follows $\alpha_r \leq \alpha_c$. The lower bound $\Gamma \geq 0 $ is trivial since we integrate a collection of Heaviside functions, which are nonnegative. We elaborated previously in sections \ref{sec:introduction}, \ref{sec:main_result}, \ref{sec:analysis} our nonconventional complex Gaussian assumption, thus we digress this argument and leave that discussion as is.

\vspace{2mm}

\begin{wrapfigure}{h}{0.42\textwidth}
  \centering
  \vspace{-1mm}
  \includegraphics[width=\linewidth]{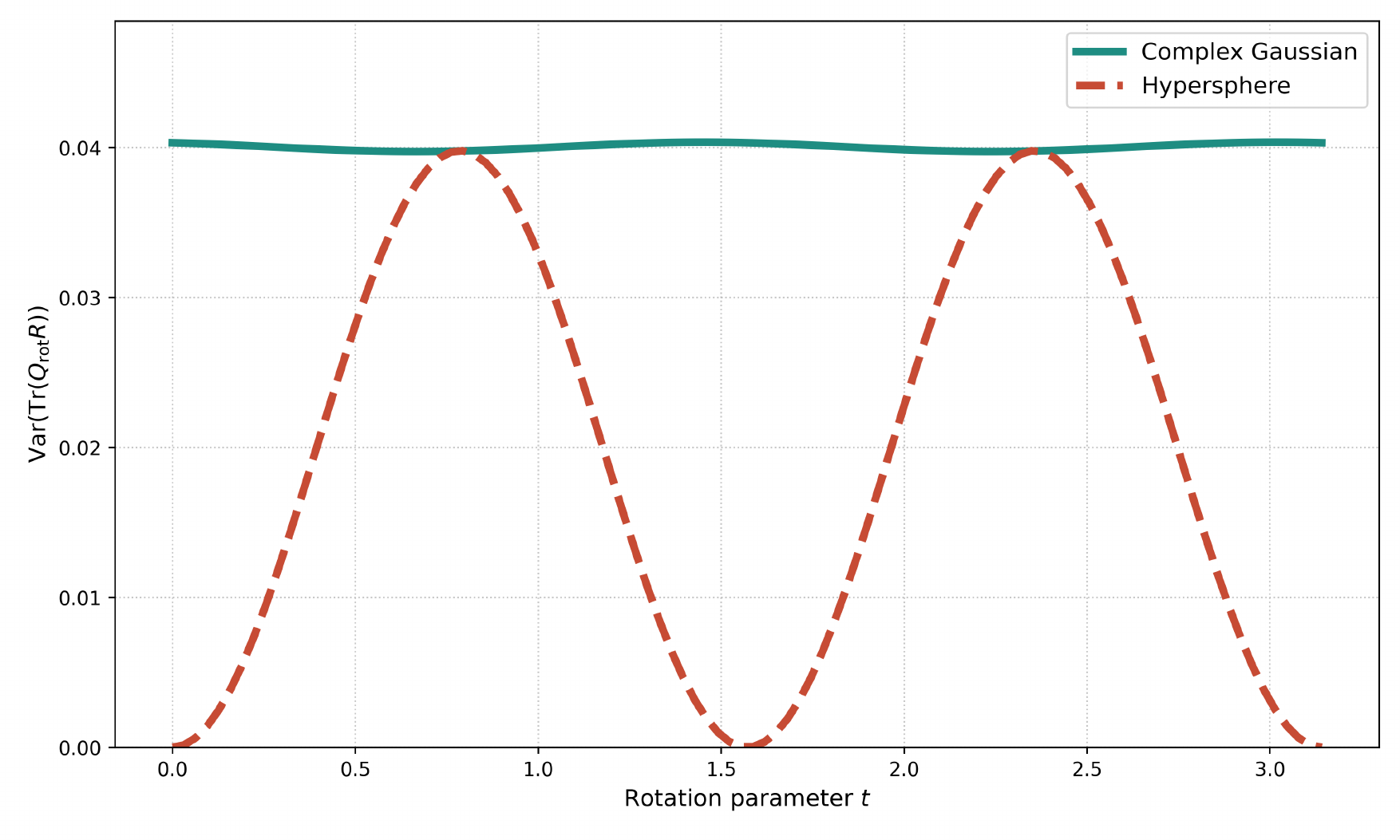}
  \vspace{-2.5mm}
  \caption{We illustrate the class invariance of the $G_0$ term as in \ref{eqn:G0} and in Appendix \ref{app:real_phase}. The straight line, corresponding to the complex Gaussian, represents the respective matrices are unchanged under a unitary transformation, parameterized by $t$, while the variance for weights along the hypersphere oscillates across a unitary transformation.}
  \vspace{-6mm}
  \label{fig:class_invariance}
\end{wrapfigure}

\textbf{Analytic justifications.} In \ref{app:proofs}, by utilizing unitary groups, hence Schur polynomials, before taking any limits, the structure of the complex version space is preserved exactly at finite $N$. When we finally invoke the discrete Laplace method, we do so over $\lambda$, rather than the continuous, infinite-dimensional space of Hermitian matrices. Evaluating an asymptotic expansion over $l_a^*$ is analytically safer and more rigorous, for example it avoids limit-swapping of traditional replica methods. In \ref{app:complex}, we downgrade from the unitary group $U(n)$ to the orthogonal group $O(n)$, and the proof is reminiscent of that in \ref{app:real_phase}. In \ref{app:complex}, the HCIZ formula fails, and we turn to the Fourier transform. In this section, we leverage the identity
\begin{align}
\mathcal{F}\Bigg[ P(\Omega) g(\Omega) \Bigg](NQ) = P\left(\frac{i}{N} \nabla_Q\right) \mathcal{F}\Bigg[ g(\Omega) \Bigg](NQ) .
\end{align}
Integration by parts then sifts the exact value out of the limit via the sifting property of Dirac masses. This provides a rigorous, exact finite-$N$ evaluation before the thermodynamic limit. 

\vspace{2mm}

\textbf{On the ansatz.} Typically, the ansatz disrupts the integration process, and is "destructive" because it places assumptions on the model before integration. An integrated ansatz is not the same as an exact integration calculation. Because the ansatz is only introduced to evaluate the asymptotic scaling of the partition lengths $\lambda_1^*$ and $\lambda_a^*$, the unitary invariance in \ref{app:real_phase} and orthogonal invariance in \ref{app:complex} of the parameter space is never broken during the integration phase.

\vspace{2mm}

\textbf{A complex-valued paradox.} Paradoxically, the real case in Appendix \ref{app:real_phase} relies on use of the unitary group, and the complex case in Appendix \ref{app:complex} relies on the orthogonal group.

\section{Conclusions and limitations}

We develop nonstandard approaches to Gardner volumes specifically for complex-valued neural networks. Our methods are analytically rigorous, as we explain in \ref{sec:discussion}, and can partially overcome pitfalls of classical methods, for example as we saw in Figure \ref{fig:ansatz} and as discussed in sections \ref{sec:introduction}, \ref{sec:discussion}. The primarily limitation of our methods as we saw in section \ref{sec:main_result} is that a closed form solution has potential to be nontrivial, but as we also saw in section \ref{sec:analysis} and Figure \ref{fig:weights_and_metric}, we can mostly overcome this.

\bibliographystyle{plainnat}
\bibliography{bibliography}

\appendix

\section{Proof of Theorem 1}

\subsection{Part I: Theorem 1, the real-phase}
\label{app:real_phase}

Construct a measure on the ambient parameter space with flat metric
\begin{align}
\label{eqn:top_form}
d\mu(\Theta) = \frac{1}{Z} e^{-K(\Theta, \overline{\Theta})} \det(h_{j\overline{k}}) \left(\frac{i}{2}\right)^N \bigwedge_{\ell=1}^N d\theta_{\ell} \wedge d\overline{\theta}_{\ell} , 
\end{align}
with Kähler potential as $K(\Theta, \overline{\Theta}) = \Theta^\dagger \Theta$, thus also $h_{j\overline{k}} = \partial_{\theta_j} \partial_{\overline{\theta}_k} K = \delta_{j\overline{k}}$. Let $\theta_j \in \mathbb{C}$ be a local coordinate representing a particular weight among the totality $\Theta = (\theta_1,\hdots,\theta_N)$, where 
\begin{align}
\label{eqn:Theta_Gaussian}
\Theta \sim \mathcal{C} \mathcal{N}(0,I_N) .
\end{align}
This setup will allow class invariance that will lead us to \ref{eqn:HCIZ}. The reason the complex Gaussian is canonically associated with the flat metric is because under the flat metric $h_{j\overline k}=\delta_{j\overline k}$, we get
\begin{align}
\|\Theta\|_h^2 = \Theta^\dagger\Theta = \sum_{j=1}^N |\theta_j|^2,
\end{align}
and $dV_h=d\Theta$, so
\begin{align}
d\mu(\Theta) = \pi^{-N}e^{-\Theta^\dagger\Theta} d\Theta ,
\end{align}
which is exactly $\mathcal{CN}(0,I_N)$.

\vspace{2mm}

Given $P$ empirical data points $Z = \{z^1, \dots, z^P\}$, let us denote $\mu$ the piece of input data, and bound the loss to prevent excess of a generalization margin $\kappa$ at point $z^{\mu}$
\begin{align}
\Gamma_{z^{\mu}}(\Theta) := - \mathcal{L}(f(z^{\mu}; \Theta), y^\mu) \ge \kappa .
\end{align}
Feeding the data into the neural network, we can notice the pre-activation is the data \cite{Gardner_1987}
\begin{align}
\label{eqn:pre-activation}
\Psi^{\mu} = \frac{1}{\sqrt{N}} \Theta^{\dagger} z^{\mu} .
\end{align}
Now, we examine a modified fractional volume of the version space via Gardner volume \cite{Shcherbina2003} \cite{sorscher2022beyond} in the complex Gaussian setting
\begin{align}
V = \int_{\mathbb{C}^N} \exp \left( - \Theta^{\dagger} \Theta \right) \frac{\omega^N}{N!} \prod_{\mu=1}^P H \Bigg(\text{Re}(\Psi^{\mu}) - \kappa \Bigg) \delta \Big( \text{Im}(\Psi^{\mu}) \Big) .
\end{align}
$\delta$, the Dirac delta, is a real slice projection. The $H$ term applied is the classic Gardner margin. We have used the definition of the Heaviside function.

\vspace{2mm}

\noindent It can be noted the complex analog is not (often) in literature. Let us denote the shorthand limit notation
\begin{align}
\lim_{( N, P, \alpha)_{\infty}} \iff \lim_{N \rightarrow \infty, P \rightarrow \infty, \frac{P}{N} = \alpha \in \mathbb{R} }
\end{align}
where $\alpha$ is a fixed real number. Let us examine the quenched free energy, or log-volume density. Using the replica identity
\begin{align}
\label{eqn:replica_limit}
f = -\lim_{( N, P, \alpha)_{\infty}} \frac{1}{N} \langle \log V \rangle = -\lim_{( N, P, \alpha)_{\infty}} \frac{1}{N} \lim_{n \to 0} \frac{\langle V^n \rangle - 1}{n}
\end{align}
since
\begin{align}
V^n = \exp(n \log V) = 1 + n \log V + \frac{(n \log V)^2}{2!} + \dots .
\end{align}
In the above, we have set $f$ which is dependent on $\langle V^n \rangle_Z$. We replicate the scenario $n$ times, yielding statistically independent parameters $\Theta^a$ for $a \in \{1,\hdots,n\}$, and we get the product
\begin{align}
V^n = \int_{(\mathbb{C}^N)^n} \left( \bigwedge_{a=1}^n \exp \left( - \Theta_a^{\dagger} \Theta_a \right) \frac{\omega_a^N}{N!} \right) \prod_{\mu=1}^P \prod_{a=1}^n H\Big(\text{Re}(\Psi_a^\mu) - \kappa\Big) \delta\Big(\text{Im}(\Psi_a^\mu)\Big) .
\end{align}
Taking the quenched average
\begin{align}
\label{eqn:quenched_avg}
\langle V^n \rangle = \int_{(\mathbb{C}^N)^n} \left( \bigwedge_{a=1}^n \exp \left( - \Theta_a^{\dagger} \Theta_a \right) \frac{\omega_a^N}{N!} \right) \left[ \left\langle \prod_{a=1}^n H\Big(\text{Re}(\Psi_a) - \kappa\Big) \delta\Big(\text{Im}(\Psi_a^\mu)\Big) \right\rangle_Z \right]^P .
\end{align}
Now it is our ambition to insert 1 via multiplication into \ref{eqn:quenched_avg}.

\vspace{2mm}

\noindent \textbf{Claim}. We have
\begin{align}
1 & = \lim_{\epsilon \to 0^+} \int_{\mathcal{H}_n} dQ \int_{i\mathcal{H}_n} d\widehat{Q}  \exp\Bigg( -N \text{Tr}\Bigg( \widehat{Q} (Q - R(\Theta)) \Bigg) \Bigg) \exp\Bigg(-\epsilon \text{Tr}(Q^2 - \widehat{Q}^2)\Bigg) ,
\end{align} 
where $Q \in \mathcal{H}_n$ is in the space of all $n\times n$ Hermitian matrices, and $\widehat{Q} \in i \mathcal{H}_n = \{ i X : X \in \mathcal{H}_n \}$, and $R(\Theta) \in \mathcal{H}_n, R_{ab}(\Theta) = \frac{1}{N} \Theta_a^{\dagger} \Theta_b$.

\vspace{2mm}

\noindent \textit{Proof of claim.} We note the standard Lebesgue measure on the space of Hermitian matrices $\mathcal{H}_n$  as
\begin{align}
dQ = \prod_{a=1}^n dQ_{aa} \prod_{a<b} d\text{Re}(Q_{ab})  d\text{Im}(Q_{ab}).
\end{align}
We will use this formula again later. For $\widehat{Q} \in i\mathcal{H}_n$, the diagonal elements are imaginary and the off-diagonal elements satisfy $\widehat{Q}_{ba} = -\overline{\widehat{Q}_{ab}}$. We define the scaled conjugate product measure on $i\mathcal{H}_n$
\begin{align}
d\widehat{Q} = \prod_{a=1}^n \frac{N}{2\pi i} d\widehat{Q}_{aa}  \prod_{a < b}  \frac{N^2}{\pi^2} d\text{Re}(\widehat{Q}_{ab})  d\text{Im}(\widehat{Q}_{ab}) .
\end{align}
Let $X = Q - R(\Theta) \in \mathcal{H}_n$. The trace inner product decomposes into $n^2$ independent real sums. Using $\widehat{Q}_{ab} = \text{Re}(\widehat{Q}_{ab}) + i\text{Im}(\widehat{Q}_{ab})$ and $X_{ab} = \text{Re}(X_{ab}) + i\text{Im}(X_{ab})$, we have:
\begin{align}
\text{Tr}(\widehat{Q}X) &= \sum_{a=1}^n \widehat{Q}_{aa} X_{aa} + \sum_{a<b} \left( \widehat{Q}_{ab} X_{ba} + \widehat{Q}_{ba} X_{ab} \right) 
\\
&= \sum_{a=1}^n \widehat{Q}_{aa} X_{aa} + 2i \sum_{a<b} \Big( \text{Im}(\widehat{Q}_{ab}) \text{Re}(X_{ab}) - \text{Re}(\widehat{Q}_{ab}) \text{Im}(X_{ab}) \Big).
\end{align}
Let us consider the integral with Gaussian regulator 
\begin{align}
\Psi_\epsilon(Q, \widehat{Q}) = \exp \{ -\epsilon \text{Tr}(Q^2) \} \exp \{\epsilon \text{Tr}(\widehat{Q}^2)\} ,
\end{align}
helping ensure integrability. Because $\widehat{Q} \in i\mathcal{H}_n$, it takes the form $\widehat{Q} = iY$ for some $Y \in \mathcal{H}_n$, so $\text{Tr}(\widehat{Q}^2) = -\text{Tr}(Y^2) \leq 0$, which yields a valid decaying Schwartz function. Taking the integral over $i\mathcal{H}_n$ with a limit, the terms factor
\begin{align}
& \lim_{\epsilon \to 0^+} \int_{i\mathcal{H}_n} d\widehat{Q} \exp\Big( -N \text{Tr}(\widehat{Q}X) + \epsilon \text{Tr}(\widehat{Q}^2) \Big) 
\\[2em]
& = \lim_{\epsilon \to 0^+} \int_{i\mathcal{H}_n} d\widehat{Q} \exp\Bigg( -N \Bigg[ \sum_{a=1}^n \widehat{Q}_{aa} X_{aa} + 2i \sum_{a<b} \Big( \text{Im}(\widehat{Q}_{ab}) \text{Re}(X_{ab})  - \text{Re}(\widehat{Q}_{ab}) \text{Im}(X_{ab}) \Big) \Bigg] + \epsilon \text{Tr}(\widehat{Q}^2) \Bigg) 
\\[2em]
\label{eqn:delta_iinfty}
& = \lim_{\epsilon \rightarrow 0^+} \prod_{a=1}^n \left[ \int_{-i\infty}^{i\infty} \exp \left\{-N \widehat{Q}_{aa} X_{aa} + \epsilon \widehat{Q}_{aa}^2 \right\} d\widehat{Q}_{aa} \right]
\\
& \ \ \ \ \ \ \ \ \ \ \ \ \times \prod_{a<b} \Bigg[ \int_{-\infty}^{\infty} \exp \left\{-2i N \text{Im}(\widehat{Q}_{ab}) \text{Re}(X_{ab}) - 2 \epsilon \text{Im}(\widehat{Q}_{ab})^2 \right\} d\text{Im}(\widehat{Q}_{ab})
\\
& \ \ \ \ \ \ \ \ \ \ \ \ \ \ \ \ \ \ \ \ \ \ \ \ \times \int_{-\infty}^{\infty} \exp \left\{2i N \text{Re}(\widehat{Q}_{ab}) \text{Im}(X_{ab}) - 2\epsilon \text{Re}(\widehat{Q}_{ab})^2  \right\} d\text{Re}(\widehat{Q}_{ab}) \Bigg] .
\end{align}
The integral splits in the above by independence, i.e. $\iint f(a) g(b) da db = (\int f(a) da)(\int g(b) db)$. Recall the Dirac delta identity
\begin{align}
\frac{1}{2\pi} \int e^{i k x} dk = \delta(x) .
\end{align}
Moreover, recall $\widehat{Q} = iY$. Thus, the integrals of \ref{eqn:delta_iinfty} evaluate to delta functions
\begin{align}
\begin{cases}
& \frac{N}{2\pi i} \int_{-\infty}^{\infty} \exp \left\{-N (i Y_{aa}) X_{aa} \right\} (i dY_{aa})  = \delta(X_{aa}) 
\\
& 
\frac{N}{\pi} \int_{-\infty}^{\infty}  \exp \left\{-2i N \text{Im}(\widehat{Q}_{ab}) \text{Re}(X_{ab})\right\} d\text{Im}(\widehat{Q}_{ab}) = \delta(\text{Re}(X_{ab}))
\\
& 
\frac{N}{\pi} \int_{-\infty}^{\infty} \exp \left\{2 i N \text{Re}(\widehat{Q}_{ab}) \text{Im}(X_{ab})\right\} d\text{Re}(\widehat{Q}_{ab}) = \delta(\text{Im}(X_{ab})).
\end{cases}
\end{align}
We have justified the evaluation of the limits via the weak-* topology of the tempered distributions, i.e. in the sense of distributions \cite{schwartz1966theorie} \cite{estrada2014support} \cite{book}. Thus, the integral decomposing to \ref{eqn:delta_iinfty} collapses to the multidimensional Dirac measure on $\mathcal{H}_n$
\begin{align}
\delta^{(n^2)}\Big(Q - R(\Theta)\Big) := \prod_{a=1}^n \delta(Q_{aa} - R_{aa}) \prod_{a<b} \delta\Big(\text{Re}(Q_{ab} - R_{ab})\Big) \delta\Big(\text{Im}(Q_{ab} - R_{ab})\Big).
\end{align}
After evaluating the limit and substituting this back into the outer integral over $Q$, we integrate a valid Dirac measure 
\begin{align}
\int_{\mathcal{H}_n} dQ  \delta^{(n^2)}\Big(Q - R(\Theta)\Big)  = 1.
\end{align}
We finish the proof. Notice
\begin{align}
& \lim_{\epsilon \rightarrow 0^+} \int_{\mathcal{H}_n} dQ \int_{i\mathcal{H}_n} d\widehat{Q} \exp\Bigg( -N \text{Tr}\Bigg( \widehat{Q} (Q - R(\Theta)) \Bigg) \Bigg) \exp\Bigg(-\epsilon \text{Tr}(Q^2) + \epsilon \text{Tr}(\widehat{Q}^2)\Bigg)
\\[1em]
& = \lim_{\epsilon_2 \to 0^+} \lim_{\epsilon_1 \to 0^+} \int_{\mathcal{H}_n} dQ \int_{i\mathcal{H}_n} d\widehat{Q} \exp\Bigg( -N \text{Tr}\Bigg( \widehat{Q} (Q - R(\Theta)) \Bigg) \Bigg) \exp\Bigg(-\epsilon_2 \text{Tr}(Q^2) + \epsilon_1 \text{Tr}(\widehat{Q}^2)\Bigg)
\\[1em]
& = \lim_{\epsilon_2 \to 0^+} \int_{\mathcal{H}_n} dQ \exp \Bigg( - \epsilon_2 \text{Tr}(Q^2) \Bigg) \left[ \lim_{\epsilon_1 \to 0^+} \int_{i\mathcal{H}_n} d\widehat{Q} \exp\Bigg( -N \text{Tr}\Bigg( \widehat{Q} (Q - R(\Theta)) \Bigg) + \epsilon_1 \text{Tr}( \widehat{Q}^2) \Bigg) \right]
\\[1em]
& = \lim_{\epsilon_2 \to 0^+} \int_{\mathcal{H}_n} dQ \exp \Bigg( - \epsilon_2 \text{Tr}(Q^2) \Bigg) \Bigg[ \delta^{(n^2)}\Big(Q - R(\Theta)\Big) \Bigg] 
\\[1em]
& = \lim_{\epsilon_2 \to 0^+} \exp \Bigg( - \epsilon_2 \text{Tr}\Big(R(\Theta)^2\Big) \Bigg) = 1.
\end{align}
It can be noted that we have used
\begin{align}
\lim_{\epsilon \rightarrow 0^+} \int f(\epsilon g + \epsilon h) d\mu = \lim_{\epsilon_1 \rightarrow 0^+} \lim_{\epsilon_2 \rightarrow 0^+} \int f(\epsilon_1 g + \epsilon_2 h)d\mu
\end{align}
under theory of distributions and suitable criteria. The above justification is similar to a Moore-Osgood criteria \cite{Loring2010}. Allow this condition (recall the hypotheses of Theorem 1).

\noindent $\square$

\vspace{2mm}

\noindent Returning to the proof via \ref{eqn:quenched_avg}, we get
\begin{align}
\label{eqn:vn}
\langle V^n \rangle & = \int_{(\mathbb{C}^N)^n} \left( \bigwedge_{a=1}^n \exp \left( - \Theta_a^{\dagger} \Theta_a \right) \frac{\omega_a^N}{N!} \right) \left[ \left\langle \prod_{a=1}^n H\Big(\text{Re}(\Psi_a^\mu) - \kappa\Big) \delta\Big(\text{Im}(\Psi_a^\mu)\Big) \right\rangle_Z \right]^P \times 1 
\\[2em]
= & \lim_{\epsilon \rightarrow 0^+} \int_{(\mathbb{C}^N)^n} \left( \bigwedge_{a=1}^n \exp \left( - \Theta_a^{\dagger} \Theta_a \right) \frac{\omega_a^N}{N!} \right) \left[ \left\langle \prod_{a=1}^n H\Big(\text{Re}(\Psi_a^\mu) - \kappa\Big) \delta\Big(\text{Im}(\Psi_a^\mu)\Big) \right\rangle_Z \right]^P 
\\
& \ \ \ \ \ \ \ \ \  \times  \int_{\mathcal{H}_n} dQ \int_{i\mathcal{H}_n} d\widehat{Q}  \exp\Bigg( -N \text{Tr}\Bigg( \widehat{Q} (Q - R(\Theta)) \Bigg) \Bigg) \exp\Bigg(-\epsilon \text{Tr}(Q^2 - \widehat{Q}^2)\Bigg)
\\[2em]
= & \lim_{\epsilon \rightarrow 0^+}  \int_{(\mathbb{C}^N)^n} \left( \bigwedge_{a=1}^n \exp \left( - \Theta_a^{\dagger} \Theta_a \right) \frac{\omega_a^N}{N!} \right) \left[ \left\langle \prod_{a=1}^n H\Big(\text{Re}(\Psi_a^\mu) - \kappa\Big) \delta\Big(\text{Im}(\Psi_a^\mu)\Big) \right\rangle_Z \right]^P 
\\
& \ \ \ \ \ \ \ \ \  \times  \int_{\mathcal{H}_n} dQ \int_{i\mathcal{H}_n} d\widehat{Q}  \exp\Bigg( - N \text{Tr}\Big( \widehat{Q} Q \Big) \Bigg) \exp \Bigg(  N \text{Tr} \Big( \widehat{Q} R(\Theta) \Big)\Bigg) \exp\Bigg(-\epsilon \text{Tr}(Q^2 - \widehat{Q}^2)\Bigg)  
\\[2em]
= & \lim_{\epsilon \rightarrow 0^+}  \int_{\mathcal{H}_n} dQ \int_{i\mathcal{H}_n} d\widehat{Q} \int_{(\mathbb{C}^N)^n} \left( \bigwedge_{a=1}^n \exp \left( - \Theta_a^{\dagger} \Theta_a \right) \frac{\omega_a^N}{N!} \right) \left[ \left\langle \prod_{a=1}^n H\Big(\text{Re}(\Psi_a^\mu) - \kappa\Big) \delta\Big(\text{Im}(\Psi_a^\mu)\Big) \right\rangle_Z \right]^P 
\\
\label{eqn:quadratic_trace}
& \ \ \ \ \ \ \ \ \  \times    \exp\Bigg( -N \text{Tr}\Big( \widehat{Q} Q \Big) \Bigg) \exp \Bigg(  N \text{Tr} \Big( \widehat{Q} R(\Theta) \Big)\Bigg) \exp\Bigg(-\epsilon \text{Tr}(Q^2 - \widehat{Q}^2)\Bigg)  .
\end{align}
The exchange of the order of integration is justified by linearity and Fubini's since the term with $\epsilon$ is a rapidly decaying function for fixed $N,\epsilon$. This term is quadratic in order of the trace in \ref{eqn:quadratic_trace}, but so is the term involving $\widehat{Q}Q$. The term involving $\widehat{Q}Q$ is oscillatory, i.e. let us take $\widehat{Q} = i\widetilde{Q}$ since $\widehat{Q}$ belongs to the class $i 
\mathcal{H}$. Into the trace, we arrive at
\begin{align}
-N \text{Tr}(\widehat{Q} Q) =- i N \text{Tr}(\widetilde{Q} Q) ,
\end{align}
and exponentiation gives a phase factor
\begin{align}
|\exp(i N \text{Tr}(\widetilde{Q} Q))| = 1 .
\end{align}
The term involving $\epsilon$ is a real damping term since $i^2 = -1$, thus it is equivalent to
\begin{align}
\exp\Big(-\epsilon \big(||Q||_F^2 + ||\widetilde{Q}||_F^2\big)\Big) .
\end{align}
Thus, Fubini's is justified. We also remark the quadratic effect in the above exponent of the term with $\epsilon$, whereas the other exponential terms are of orders belonging to $\widehat{Q}Q,\widehat{Q}R(\Theta)$ in their exponents.

\vspace{2mm}

\noindent We can see after the substitution
\begin{align}
\label{eqn:H_to_exp}
\left[ \left\langle \prod_{a=1}^n H\Big(\text{Re}(\Psi_a^\mu) - \kappa\Big) \delta\Big(\text{Im}(\Psi_a^\mu)\Big) \right\rangle_Z \right]^P \implies \exp \Bigg( P G_1(Q) \Bigg) .
\end{align}
The above of \ref{eqn:H_to_exp} is nontrivial so we elaborate. The left-hand side is a function of $\Theta$ while the right-hand side is a function of $Q$. From the claim, it follows
\begin{align}
\langle V^n \rangle & = \int_{(\mathbb{C}^N)^n} \left( \bigwedge_{a=1}^n \exp \left( - \Theta_a^{\dagger} \Theta_a \right) \frac{\omega_a^N}{N!} \right) \int_{\mathcal{H}_n} dQ \ \delta^{(n^2)}\Big(Q - R(\Theta)\Big) \left[ \left\langle \prod_{a=1}^n H\Big(\text{Re}(\Psi_a^\mu) - \kappa\Big) \delta\Big(\text{Im}(\Psi_a^\mu)\Big) \right\rangle_Z \right]^P
\\[2em]
& = \int_{(\mathbb{C}^N)^n} \left( \bigwedge_{a=1}^n \exp \left( - \Theta_a^{\dagger} \Theta_a \right) \frac{\omega_a^N}{N!} \right) \int_{\mathcal{H}_n} dQ \ \delta^{(n^2)}\Big(Q - R(\Theta)\Big) \exp \Bigg( P G_1(Q) \Bigg)
\\[2em]
\label{eqn:G0}
& =\lim_{\epsilon \rightarrow 0^+} \int_{(\mathbb{C}^N)^n} \left( \bigwedge_{a=1}^n \exp \left( - \Theta_a^{\dagger} \Theta_a \right) \frac{\omega_a^N}{N!} \right) \int_{\mathcal{H}_n} dQ \int_{i\mathcal{H}_n} d\widehat{Q} \exp \Bigg( P G_1(Q) \Bigg)
\\
 & \ \ \ \ \ \ \ \ \ \ \ \ \ \ \ \ \ \  \times \exp\Bigg( -N \text{Tr}\Bigg( \widehat{Q} (Q - R(\Theta)) \Bigg) \Bigg) \exp\Bigg(-\epsilon \text{Tr}(Q^2 - \widehat{Q}^2)\Bigg) .
\end{align}
Therefore, we can write $\langle V^n \rangle$ with two different forms via the claim. The claim implies that there is a fundamental dependence between $Q, R(\Theta)$ since the Dirac delta is a restriction that enforces a relation between the two, i.e. under the Dirac measure forces an identity relationship. This is nontrivial so we elaborate more. The Dirac delta $\delta(Q - R(\Theta))$ tells us that $Q = R(\Theta)$. It does not give us permission to take any function $f(\Theta)$ and rewrite it as $f(Q)$. 

\vspace{2mm}

Fortunately, we can achieve the desired effect because of the nature of the Gaussian measure. For any given pattern $z^\mu \sim \mathcal{C}\mathcal{N}(0, I_N)$, we define the pre-activation vector across all $n$ replicas as $\Psi^\mu = (\Psi_1^\mu, \dots, \Psi_n^\mu)^T$. By definition, as before in \ref{eqn:pre-activation}
\begin{align}
\Psi^\mu = \frac{1}{\sqrt{N}} \Theta^\dagger z^\mu .
\end{align}
Because linear combinations of centered Gaussian random variables are themselves Gaussian, $\Psi^\mu$ is a complex Gaussian random vector in $\mathbb{C}^n$ with zero mean. We examine the $n \times n$ covariance matrix, which evaluates to
\begin{align}
\mathbb{E}_Z \left[ \Psi^\mu (\Psi^\mu)^\dagger \right] &= \frac{1}{N} \Theta^\dagger \mathbb{E}_Z \left[ z^\mu (z^\mu)^\dagger \right] \Theta  \\
&= \frac{1}{N} \Theta^\dagger I_N \Theta  = R(\Theta) .
\end{align}
Equivalently, $\Psi^{\mu} | \Theta \sim \mathcal{CN}(0, R(\Theta))$. Hence
\begin{align}
\left\langle \prod_{a=1}^n H\Big(\text{Re}(\Psi_a^\mu) - \kappa\Big) \delta\Big(\text{Im}(\Psi_a^\mu)\Big) \right\rangle_Z \Bigg|_{\Theta} = 
\left\langle \prod_{a=1}^n H\Big(\text{Re}(\Psi_a^\mu) - \kappa\Big) \delta\Big(\text{Im}(\Psi_a^\mu)\Big) \right\rangle_Z \Bigg|_{\Theta'}.
\end{align}
That means the expectation is constant on the fibers of the map $
\Theta \mapsto R(\Theta)$.
This validates our substitution, i.e.
\begin{align}
\label{eqn:G1}
G_1(Q) =  \log \left\langle \prod_{a=1}^n H\Big(\text{Re}(\Psi_a^\mu) - \kappa\Big) \delta\Big(\text{Im}(\Psi_a^\mu)\Big) \right\rangle_Z  \Bigg|_{R(\Theta) = Q} .
\end{align}
We provide more basic discussion in Appendix \ref{app:discussion}. Also, we define
\begin{align}
\exp\Big(N G_0(\widehat{Q})\Big) := \int_{(\mathbb{C}^N)^n} \left( \bigwedge_{a=1}^n \exp \left( - \Theta_a^{\dagger} \Theta_a \right) \frac{\omega_a^N}{N!} \right) \exp \Bigg( N \text{Tr} \Big( \widehat{Q} R(\Theta) \Big)\Bigg) ,
\end{align}
which we also mention in \ref{sec:discussion}.

\vspace{2mm}

\noindent Since $Q$ is Hermitian and $\widehat{Q}$ is anti-Hermitian, they admit spectral decompositions
\begin{align}
Q = U \Lambda U^\dagger, \quad \widehat{Q} = \widehat{U} \widehat{\Lambda} \widehat{U}^\dagger .
\end{align}
$U, \widehat{U}$ are unitary, and $\Lambda = \text{diag}(\lambda_1,\hdots,\lambda_n), \widehat{\Lambda} = \text{diag}(\widehat{\lambda}_1,\hdots,\widehat{\lambda}_n)$ are diagonal matrices of eigenvalues. Via the Weyl integration formula, we get a change of measure
\begin{align}
dQ = \left( \frac{\pi^{n(n-1)/2}}{\prod_{j=1}^n j!} \right) \Delta(\Lambda)^2 d\Lambda dU, \quad d\widehat{Q} = (-1)^{\frac{n(n-1)}{2}} \left(  \frac{\pi^{n(n-1)/2}}{\prod_{j=1}^n j!} \right) \Delta(\widehat{\Lambda})^2 d\widehat{\Lambda} d\widehat{U} .
\end{align}
Here, $\Delta(\chi)^2$ is the Vandermonde determinant $\Delta(\chi)^2 = \prod_{1 \leq a < b \leq n} (\lambda_{\chi,b} - \lambda_{\chi,a})^2$. $dU, d\widehat{U}$ are normalized Haar measures on the unitary group. Without loss of generality, the $(-1)^{\frac{n(n-1)}{2}}$ signed term will cancel the signed term from use of the HCIZ formula later in the proof, so we will omit the sign in our subsequent computations. We have used the fact that $dQ$ is the volume product of differentials
\begin{align}
dQ = \prod_{i} dQ_{ii} \prod_{1 \leq i < j \leq n} d(\text{Re}(Q_{ij})) d(\text{Im}(Q_{ij})) = \frac{\pi^{n(n-1)/2}}{\prod_{j=1}^n j!} \prod_{1 \leq i < j \leq n} (\lambda_i - \lambda_j)^2 \left( d\lambda_1 \dots d\lambda_n \right) dU .
\end{align}
Using the cyclic property of the trace,
\begin{align}
\text{Tr}(Q\widehat{Q}) = \text{Tr}(U \Lambda U^\dagger \widehat{U} \widehat{\Lambda} \widehat{U}^\dagger) = \text{Tr}(\Lambda V \widehat{\Lambda} V^\dagger) .
\end{align}
We have set $V = U^{\dagger} \widehat{U}$. Keeping track of constants, the final constant used in the integration is
\begin{align}
\left( \frac{\pi^{n(n-1)}}{\left( \prod_{j=1}^n j! \right)^2} \right) \left( \frac{N^{n^2}}{2^n \pi^{n^2} i^n} \right) = \frac{N^{n^2}}{i^n 2^n \pi^n \left( \prod_{j=1}^n j! \right)^2} .
\end{align}
Applying our decomposition to \ref{eqn:vn}, we get a rewriting
\begin{align}
\label{eqn:real_Vn_decomposed}
\langle V^n \rangle & = \left( \frac{1}{\left( \prod_{j=1}^n j! \right)^2} \frac{N^{n^2}}{i^n 2^n \pi^n} \right) \lim_{\epsilon \rightarrow 0^+} \int_{\mathbb{R}^n} d\Lambda \int_{(i\mathbb{R})^n} d\widehat{\Lambda} \Delta(\Lambda)^2 \Delta(\widehat{\Lambda})^2
\\
& \ \ \ \ \ \ \ \ \ \times \exp \Bigg( N G_0(\widehat{\Lambda}) \Bigg) \exp\Bigg(-\epsilon \text{Tr}(\Lambda^2 - \widehat{\Lambda}^2)\Bigg)
\\
& \ \ \ \ \ \ \ \ \ \times \vast[ \int_{U(n)} d\widehat{U} \int_{U(n)} dU  \exp \Bigg( -N \text{Tr}\Big(\Lambda (U^\dagger \widehat{U}) \widehat{\Lambda} (U^\dagger \widehat{U})^\dagger \Big) \Bigg) \exp \Bigg( P G_1(U \Lambda U^\dagger) \Bigg) \vast]
\\[2em]
& = \left( \frac{1}{\left( \prod_{j=1}^n j! \right)^2} \frac{N^{n^2}}{i^n 2^n \pi^n} \right) \lim_{\epsilon \rightarrow 0^+} \int_{\mathbb{R}^n} d\Lambda \int_{(i\mathbb{R})^n} d\widehat{\Lambda} \Delta(\Lambda)^2 \Delta(\widehat{\Lambda})^2
\\
& \ \ \ \ \ \ \ \ \ \times \exp \Bigg( N G_0(\widehat{\Lambda}) \Bigg) \exp\Bigg(-\epsilon \text{Tr}(\Lambda^2 - \widehat{\Lambda}^2)\Bigg)
\\
& \ \ \ \ \ \ \ \ \ \times \vast[ \int_{U(n)} dV \exp \Bigg( -N \text{Tr}\Big(\Lambda V \widehat{\Lambda} V^\dagger \Big) \Bigg) \vast] \vast[ \int_{U(n)} dU \exp \Bigg( P G_1(U \Lambda U^\dagger) \Bigg) \vast] .
\end{align}
The exchange of the unitary integrals is by linearity but also consider Fubini's \cite{Ziemer2017}. It can be noted the unitary group $U(n)$ is compact, and our functions are sufficiently nice.

\vspace{2mm}

\noindent We will use the following fact: any class function on the unitary group (or invariant function on Hermitian matrices) can be uniquely expanded in terms of Schur polynomials $s_\lambda(Q)$ \cite{Weyl1940} \cite{Goodman2009}. We get $\exp(c_{N,P} G(A)) = \sum_{\lambda} c_\lambda(P) s_\lambda(A)$ \cite{Balantekin_2000} \cite{Schlittgen_2003} when $G$ is a class function. 

\vspace{2mm}

$G_0$ is automatically a class function due to the complex Gaussian weights. We refer to Figure \ref{fig:class_invariance} for empirical evidence. It can be noted that $G_1$ is not a class function due to the asymmetric real and imaginary constraints. We cannot decompose it into Schur polynomials immediately. The integral $\int_{U(n)} dU \exp(P G_1(U \Lambda U^\dagger))$ acts as an averaging mechanism over all complex phases. We define a new function $\widetilde{G}_1(\Lambda)$ based on \ref{eqn:real_Vn_decomposed} such that
\begin{align}
\label{eqn:U_invariance}
\exp \Bigg( P \widetilde{G}_1(\Lambda) \Bigg) := \int_{U(n)} dU \exp \Bigg( P G_1(U \Lambda U^\dagger) \Bigg) 
\end{align}
(which is well-defined via a log). Now $G_0(\widehat{\Lambda})$ and $\widetilde{G}_1(\Lambda)$ are both symmetric class functions, and we can decompose them with Schur polynomials.

\vspace{2mm}

\noindent Now, consider our change of variables on $\widehat{U}$. Since $V = U^{\dagger} \widehat{U}$ and by unitary matrix properties, we get $\widehat{U} = UV$. Because the Haar measure on the unitary group is left-invariant, we get $d\widehat{U} = dV$. This justifies the two inner integrals becoming
\begin{align}
\vast[ \int_{U(n)} dV \exp \Bigg( -N \text{Tr}\Big(\Lambda V \widehat{\Lambda} V^\dagger \Big) \Bigg) \vast] \vast[ \int_{U(n)} dU \exp \Bigg( P G_1(U \Lambda U^\dagger) \Bigg) \vast] .
\end{align}
The inner integral of the above evaluates to the Harish-Chandra-Itzykson-Zuber formula \cite{Tao2013HCIZ}
\begin{align}
\label{eqn:HCIZ}
\int_{U(n)} dV \exp \Bigg( -N \text{Tr}\Big(\Lambda V \widehat{\Lambda} V^\dagger \Big) \Bigg) = \left( \prod_{j=1}^{n-1} j! \right) \frac{(-1)^{\frac{n(n-1)}{2}}}{N^{\frac{n(n-1)}{2}}} \frac{\det\left( \exp \{ -N \lambda_a \widehat{\lambda}_b \}\right)_{a,b=1}^n}{\Delta(\Lambda) \Delta(\widehat{\Lambda})} .
\end{align}
Therefore, the partition function of \ref{eqn:real_Vn_decomposed} collapses, again omitting the sign from the change of measures earlier,
\begin{align}
\langle V^n \rangle & = \left( \frac{ N^{\frac{n(n+1)}{2}}}{i^n (2\pi)^n n! \prod_{j=1}^n j!} \right) \lim_{\epsilon \rightarrow 0^+}  \int_{\mathbb{R}^n} d\Lambda \int_{(i\mathbb{R})^n} d\widehat{\Lambda} \Delta(\Lambda) \Delta(\widehat{\Lambda}) \sum_{\lambda} \sum_{\mu} c_{\lambda}(P) c_{\mu}(N) s_{\mu}(\widehat{\Lambda})  s_{\lambda}(\Lambda)
\\
& \ \ \ \ \ \ \ \ \times \det\Bigg(\exp \Bigg( {-N \lambda_a \widehat{\lambda}_b} \Bigg) \Bigg) \exp\Bigg(-\epsilon \text{Tr}(\Lambda^2 - \widehat{\Lambda}^2)\Bigg).
\end{align}
It can be noted one power of $\Delta ( \chi)$ has been annihilated. Let us pull the infinite sum out of the integration. 

\vspace{2mm}

\noindent Now we invoke Fubini's theorem, or the discrete dominated convergence on the counting measure, for fixed $\epsilon > 0$ to pull the sums out of the integral
\begin{align}
\label{eqn:sums_outside_int}
& \langle V^n \rangle = \left( \frac{(-1)^{\frac{n(n-1)}{2}} N^{\frac{n(n+1)}{2}}}{i^n (2\pi)^n n! \prod_{j=1}^n j!} \right) \lim_{\epsilon \to 0^+} \sum_{\lambda} \sum_{\mu} c_{\lambda}(P) c_{\mu}(N)    \int_{\mathbb{R}^n} d\Lambda \int_{(i\mathbb{R})^n} d\widehat{\Lambda} \Delta(\Lambda) \Delta(\widehat{\Lambda})  s_{\mu}(\widehat{\Lambda})  s_{\lambda}(\Lambda)
\\
& \ \ \ \ \ \ \ \ \ \ \ \ \ \ \ \ \ \ \ \ \ \ \ \ \ \ \ \ \ \  \times \det\Bigg(\exp \{-N \lambda_a \widehat{\lambda}_b\}\Bigg)  \exp\Bigg(-\epsilon \text{Tr}(\Lambda^2 - \widehat{\Lambda}^2)\Bigg)   .
\end{align}

\vspace{2mm}

\noindent Let us define the regularized integral
\begin{align}
I_{\lambda, \mu}(\epsilon) = \int_{\mathbb{R}^n} d\Lambda \int_{(i\mathbb{R})^n} d\widehat{\Lambda} \Delta(\Lambda) \Delta(\widehat{\Lambda}) s_{\mu}(\widehat{\Lambda})  s_{\lambda}(\Lambda)  \det\Bigg(\exp \{-N \lambda_a \widehat{\lambda}_b\}\Bigg) \exp\Bigg(-\epsilon \text{Tr}(\Lambda^2 - \widehat{\Lambda}^2)\Bigg)  .
\end{align}
Denote partitions $\lambda = (\lambda_1, \dots, \lambda_n)$, $\mu = (\mu_1, \dots, \mu_n)$. We define the decreasing shifted parts as $l_a = \lambda_a + n - a, m_b = \mu_b + n - b$. The Schur polynomials rewrite exactly as \cite{macdonald1995symmetric}
\begin{align}
& s_\lambda(\Lambda) = \frac{\det(\lambda_a^{l_b})}{\Delta(\Lambda)} \implies \Delta(\Lambda) s_\lambda(\Lambda) = \det(\lambda_a^{l_b})
\\
& s_\mu(\widehat{\Lambda}) = \frac{\det(\widehat{\lambda}_c^{m_d})}{\Delta(\widehat{\Lambda})} \implies \Delta(\widehat{\Lambda}) s_\mu(\widehat{\Lambda}) = \det(\widehat{\lambda}_c^{m_d}) .
\end{align}
Here, $\lambda_{\gamma}^{l_i}$ is an element of the alternant matrix
\begin{align}
\det(\lambda_{\gamma}^{l_b}) = \det \begin{pmatrix}
\lambda_1^{l_1} & \lambda_1^{l_2} & \cdots & \lambda_1^{l_n} \\
\lambda_2^{l_1} & \lambda_2^{l_2} & \cdots & \lambda_2^{l_n} \\
\vdots & \vdots & \ddots & \vdots \\
\lambda_n^{l_1} & \lambda_n^{l_2} & \cdots & \lambda_n^{l_n}
\end{pmatrix} ,
\end{align}
which is a Vandermonde matrix under the selected powers. Therefore, the integral of interest becomes
\begin{align}
\label{eqn:I_int}
I_{\lambda, \mu}(\epsilon) = \int_{\mathbb{R}^n} d\Lambda \int_{(i\mathbb{R})^n} d\widehat{\Lambda} \det(\lambda_a^{l_b}) \det(\widehat{\lambda}_c^{m_d}) \det\Bigg(\exp \{-N \lambda_j \widehat{\lambda}_k \}\Bigg) \exp\Bigg(-\epsilon \text{Tr}(\Lambda^2 - \widehat{\Lambda}^2)\Bigg)   .
\end{align}
The generalized Andreief identity states (with respect to Lebesgue measure) \cite{forrester2018meetandreiefbordeaux1886}
\begin{align}
\frac{1}{(n!)^2} \iint \det(f_a(x_b)) \det(g_c(y_d)) \det(K(x_j, y_k)) dx dy = \det \left( \iint f_a(x) g_d(y) K(x,y) dx dy \right) .
\end{align}
where $K : X \times Y \rightarrow \mathbb{C}$ is a kernel and $f : X \rightarrow \mathbb{C},g : Y \rightarrow \mathbb{C}$, $\det(h_i(z_j))$ is the matrix determinant indexed by function $i$ and point $j$, and $K(x_j,y_k)$ is shorthand for $K$ evaluated at all pairs $(x_j,y_k)$.
Also, we note in \ref{eqn:I_int}
\begin{align}
\exp\Bigg(-\epsilon \text{Tr}(\Lambda^2 - \widehat{\Lambda}^2)\Bigg) = \prod_{j=1}^n \exp(-\epsilon \lambda_j^2) \prod_{k=1}^n \exp(\epsilon \widehat{\lambda}_k^2) .
\end{align}
Therefore, our integral of \ref{eqn:I_int} collapses to
\begin{align}
\label{eqn:real_Phi}
I_{\lambda, \mu}(\epsilon) = (n!)^2 \det(\Phi_{ad}(\epsilon)),  \quad \Phi_{ad}(\epsilon) = \iint d\lambda d\widehat{\lambda} \lambda^{l_a} \widehat{\lambda}^{m_d} \exp \{-N \lambda \widehat{\lambda}  - \epsilon(  \lambda^2  - \widehat{\lambda}^2 ) \}  .
\end{align}
Now we evaluate the matrix element $\Phi$. Recall $Q \in \mathcal{H}_n$, and so $\lambda$ is real, and the conjugate variable $\widehat{Q} \in i\mathcal{H}_n$, meaning $\widehat{\lambda}$ is purely imaginary. We use notation $\widehat{\lambda} = i \Omega$. It follows from \ref{eqn:real_Phi}
\begin{align}
\label{eqn:phi_integral}
\Phi_{ad}(\epsilon) & = i \int_{-\infty}^\infty d\lambda \int_{-\infty}^\infty d\Omega  \lambda^{l_a} (i\Omega)^{m_d} \exp \{-i N \lambda \Omega \} \exp\{ - \epsilon ( \lambda^2  + \Omega^2) \} 
\\
& =  i \int_{-\infty}^\infty d\lambda  \lambda^{l_a} \exp\{ - \epsilon  \lambda^2 \} \int_{-\infty}^\infty d\Omega  (i\Omega)^{m_d} \exp \{-i N \lambda \Omega - \epsilon \Omega^2 \}  .
\end{align}
Observe the inner integral is a Fourier transform. Therefore, differentiating in Fourier space,
\begin{align}
\int_{-\infty}^\infty  d\Omega (i\Omega)^{m_d} \exp \{-\epsilon \Omega^2 \} \exp \{-i N \lambda \Omega \}  = (-1)^{m_d} \frac{d^{m_d}}{d(N\lambda)^{m_d}} \left( \sqrt{\frac{\pi}{\epsilon}} \exp\left\{ -\frac{N^2 \lambda^2}{4\epsilon} \right\} \right) .
\end{align}
Therefore, substituting back into \ref{eqn:phi_integral}
\begin{align}
\Phi_{ad}(\epsilon) & = i \int_{-\infty}^\infty d\lambda \lambda^{l_a} \exp(-\epsilon \lambda^2) \left[ (-1)^{m_d} \frac{d^{m_d}}{d(N\lambda)^{m_d}} \left( \sqrt{\frac{\pi}{\epsilon}} \exp\left\{ -\frac{N^2 \lambda^2}{4\epsilon} \right\} \right) \right] 
\\[2em]
& =  \frac{i (-1)^{m_d}}{N^{m_d}} \sqrt{\frac{\pi}{\epsilon}} \int_{-\infty}^\infty d\lambda \lambda^{l_a} \exp(-\epsilon \lambda^2) \frac{d^{m_d}}{d\lambda^{m_d}} \left( \exp\left\{ -\frac{N^2 \lambda^2}{4\epsilon} \right\} \right)
\\[2em]
& = \frac{i (-1)^{2m_d}}{N^{m_d}} \sqrt{\frac{\pi}{\epsilon}} \int_{-\infty}^\infty d\lambda \left[ \frac{d^{m_d}}{d\lambda^{m_d}} \left( \lambda^{l_a} \exp(-\epsilon \lambda^2) \right) \right] \exp\left\{ -\frac{N^2 \lambda^2}{4\epsilon} \right\}
\\[2em]
&  = \frac{i}{N^{m_d}} \int_{-\infty}^\infty d\lambda \left[ \frac{d^{m_d}}{d\lambda^{m_d}} \left( \lambda^{l_a} \exp(-\epsilon \lambda^2) \right) \right] \left( \sqrt{\frac{\pi}{\epsilon}} \exp\left\{ -\frac{N^2 \lambda^2}{4\epsilon} \right\} \right) .
\end{align}
The third line follows by integration by parts with decay. The term in the parentheses is the nascent Dirac delta function \cite{sun2025fastergodicsearchkernel}, which scales 
\begin{align}
\lim_{\epsilon \to 0^+} \sqrt{\frac{\pi}{\epsilon}} \exp\left\{ -\frac{N^2 \lambda^2}{4\epsilon} \right\} = \frac{2\pi}{N} \delta(\lambda) .
\end{align}
This collapses the integral, yielding in regard to \ref{eqn:real_Phi}
\begin{align}
\lim_{\epsilon \to 0^+} \Phi_{ad}(\epsilon) = \frac{i}{N^{m_d}} \frac{2\pi}{N} \left[ \frac{d^{m_d}}{d\lambda^{m_d}} \left( \lambda^{l_a} \exp(-\epsilon \lambda^2) \right) \right]_{\lambda=0} .
\end{align}
We can remark the above is justified rigorously via
\begin{align}
\lim_{\epsilon \to 0^+} \int_{-\infty}^\infty f_\epsilon(\lambda) \delta_\epsilon(\lambda) d\lambda  ,
\end{align}
and follows from the theory of distributions similar to how the claim, discussed in \ref{sec:subsidiary}, was derived. The only nonzero terms come from $l_a = m_d$, thus
\begin{align}
\lim_{\epsilon \to 0^+} \Phi_{ad}(\epsilon) = \frac{2\pi i}{N^{m_d+1}} l_a! \delta_{l_a, m_d} .
\end{align}
We can note
\begin{align}
\det(\Phi) = \frac{(2\pi i)^n}{N^{|\lambda| + \frac{n(n+1)}{2}}} \left( \prod_{a=1}^n l_a! \right) \delta_{\lambda, \mu} ,
\end{align}
noting constant contribution.

\vspace{2mm}

\noindent Now, we return to $\langle V^n \rangle$. This means the integral \ref{eqn:I_int} collapses into a single sum
\begin{align}
\langle V^n \rangle = \sum_{\lambda} \sum_{\mu} c_\lambda(P)c_\mu(N) \Bigg[ \left( \frac{N^{\frac{n(n+1)}{2}}}{i^n (2\pi)^n n! \prod_{j=1}^n j!} \right) (n!)^2 \frac{(2\pi i)^n}{N^{|\lambda| + \frac{n(n+1)}{2}}} \left( \prod_{a=1}^n l_a! \right) \delta_{\lambda, \mu} \Bigg],
\end{align}
which simplifies to
\begin{align}
\langle V^n \rangle = \left( \frac{n!}{\prod_{j=1}^n j!} \right) \sum_{\lambda} c_{\lambda}(P) c_{\lambda}(N) \frac{\prod_{a=1}^n l_a!}{N^{|\lambda|}}   .
\end{align}
Recall the identity $x = \exp \log x$. It subsequently follows
\begin{align}
\label{eqn:real_Vn_exp}
\langle V^n \rangle & = \sum_{\lambda} \exp \Bigg\{ \log \Bigg\{ \left( \frac{n!}{\prod_{j=1}^n j!} \right) c_{\lambda}(P) c_{\lambda}(N) N^{\frac{n(n-1)}{2}} \left( \prod_{a=1}^n \frac{l_a!}{N^{l_a}} \right) \Bigg\} \Bigg\}
\\
& = \sum_{\lambda} \exp \Bigg\{  \log \left( \frac{n!}{\prod_{j=1}^n j!} \right) + \log c_{\lambda}(P) + \log c_{\lambda}(N) + \frac{n(n-1)}{2} \log N + \sum_{a=1}^n \Big( l_a \log l_a - l_a - l_a \log N \Big)  \Bigg\} 
\\
& := \sum_{\lambda} \exp \{  S(\lambda,n) \},
\end{align}
which is an application of log properties, and the product becomes a sum by Stirling's approximation. We have also noted
\begin{align}
\frac{1}{N^{|\lambda|}} = \frac{N^{\frac{n(n-1)}{2}}}{N^{\sum l_a}} = N^{\frac{n(n-1)}{2}} \prod_{a=1}^n \frac{1}{N^{l_a}}  .
\end{align}
By a discrete version of Laplace's method on \ref{eqn:real_Vn_exp},
\begin{align}
\label{eqn:real_laplace}
\sum_{\lambda} \exp \Big\{ S(\lambda, n) \Big\} \xrightarrow{N \to \infty} \left( \frac{1}{\Delta x} \right)^n \underbrace{ \sqrt{\frac{(2\pi)^n}{\det \Big( - \mathcal{H}_{\lambda^*}(S) \Big)}} }_{=C(n)} \exp \Big\{ \text{extr}_{\lambda^*} S(\lambda^*, n) \Big\} ,
\end{align}
where the saddle point partition $\lambda^*$ satisfies the extremal condition $\frac{\partial S}{\partial l_a} = 0$. Note that we are not technically evaluating the limit yet, but rather it is an asymptotic expansion. $\mathcal{H} $ is a corresponding Hessian term. It follows using \ref{eqn:real_laplace}
\begin{align}
f=-\lim_{( N, P, \alpha)_{\infty}} \frac{1}{N} \lim_{n \to 0} \frac{\log \langle V^n \rangle}{n} = -\lim_{( N, P, \alpha)_{\infty}} \frac{1}{N} \left( \left. \frac{\partial S(\lambda^*, n)}{\partial n} \right|_{n=0} + \left. \frac{\partial \log C(n)}{\partial n} \right|_{n=0} \right) .
\end{align}
This follows from the formal definition of the partial derivative, and $S(\lambda^*, 0) = 0$. To evaluate this derivative, we transition to the thermodynamic limit $N \to \infty$. At critical capacity,
\begin{align}
-\lim_{( N, P, \alpha)_{\infty}} \frac{1}{N} \left( \left. \frac{\partial S(\lambda^*, n)}{\partial n} \right|_{n=0} + \left. \frac{\partial \log C(n)}{\partial n} \right|_{n=0} \right) = 0 .
\end{align}
By linearity,
\begin{align}
\label{eqn:real_f}
& \underbrace{ \frac{\partial}{\partial n} \log \left( \frac{n!}{\prod_{j=1}^n j!} \right) \Bigg|_{n=0} }_{\mathcal{S}_{\text{constant}}}
+ \underbrace{ \frac{\partial \log c_{\lambda^*}(N)}{\partial n} \Bigg|_{n=0} }_{\mathcal{S}_{\text{prior}}} \nonumber \\
& + \underbrace{ \frac{\partial}{\partial n} \sum_{a=1}^n (l_a^* \log l_a^* - l_a^*) \Bigg|_{n=0} }_{\mathcal{S}_{\text{mixing}}}
+ \underbrace{ \frac{\partial}{\partial n} \left( \sum_{a=1}^n l_a^* \log N - \frac{n(n-1)}{2} \log N \right) \Bigg|_{n=0} }_{\mathcal{S}_{\text{scaling}}} \nonumber \\
& + \underbrace{ \frac{\partial \log c_{\lambda^*}(P)}{\partial n} \Bigg|_{n=0} }_{\mathcal{S}_{\text{patterns}}(\alpha_r)}
+ \underbrace{ \frac{\partial \log C(n)}{\partial n} \Bigg|_{n=0} }_{\mathcal{S}_{\text{fluctuations}}} = f .
\end{align}
Recall we had set $l_a = \lambda_a + n - a$. By definition, a valid integer partition requires $\lambda_1 \geq \lambda_2 \geq \dots \geq \lambda_n$. Under the replica symmetric ansatz, the overlap matrix $Q$ has $Q_{aa} = q_d$ and $Q_{ab} = q$. The diagonal $Q_{aa}$ is no longer 1 due to the complex Gaussian of \ref{eqn:Theta_Gaussian}. The eigenvalues here are $q_d + (n-1)q$ and $q_d - q$ with algebraic and geometric multiplicity $n-1$. Therefore, we can take in large $N$ limit $\lambda_1^* = N(q_d + (n-1)q)$ and $\lambda_a^* = N(q_d - q)$ for $a = 2, \dots, n$. This yields $l_1^* = N(q_d + (n-1)q) + n - 1$ and $l_a^* = N(q_d - q) + n - a$ for $a \geq 2$. This preserves the inequality $l_1^* > l_2^* > \dots > l_n^*$.

\vspace{2mm}

Let us consider large $N$ asymptotics. Notice
\begin{align}
\sum_{a=1}^n l_a^*
= (N q_1 + n - 1) + \sum_{a=2}^n (N q_0 + n - a)
= N \big(q_d + (n-1)q + (n-1)(q_d-q)\big) + \frac{n(n-1)}{2}
= n N q_d + \frac{n(n-1)}{2} .
\end{align}
Substituting this into the argument for the scaling term in \ref{eqn:real_f},
\begin{align}
\label{eqn:scaling_term}
\sum_{a=1}^n l_a^* \log N - \frac{n(n-1)}{2} \log N
= n N q_d \log N + \frac{n(n-1)}{2} \log N - \frac{n(n-1)}{2} \log N
= n N q_d \log N .
\end{align}
Thus, evaluating the linear derivative at the replica limit $n \to 0$ for the scaling term of \ref{eqn:scaling_term} gives
\begin{align}
\frac{\partial}{\partial n} \Big[ n N q_d \log N \Big] \Big|_{n=0} = N q_d \log N .
\end{align}
For the mixing term in \ref{eqn:real_f}, expanding to leading order in $N$ yields $\sum_{a=1}^n (l_a^* \log l_a^* - l_a^*) \approx N q_1 \log(N q_1) - N q_1 + (n-1) \big( N q_0 \log(N q_0) - N q_0 \big)$. Taking the derivative at $n=0$
\begin{align}
\left.
\frac{\partial}{\partial n}
\sum_{a=1}^n
\big(l_a^*\log l_a^*-l_a^*\big)
\right|_{n=0}
=
Nq_d\log\big(N(q_d-q)\big)
-N(q_d-q).
\end{align}
Now, we apply the $\frac{1}{N}$ scaling and transition to the thermodynamic limit
$N \to \infty$. Combining the mixing and scaling evaluations
\begin{align}
&\lim_{N\to\infty}\frac{1}{N} \Big[ Nq_d\log\big(N(q_d-q)\big)  N(q_d-q) -Nq_d\log N \Big]
\\
&\qquad 
= q_d\log(q_d-q)-q_d+q.
\end{align}
The sub-exponential fluctuation term and the constant term vanish in this limit since $N$ scales faster than their bounds
\begin{align}
\lim_{N \to \infty} \frac{1}{N} \mathcal{S}_{\text{fluctuations}} = 0 , \qquad
\lim_{N \to \infty} \frac{1}{N} \mathcal{S}_{\text{constant}} = 0 .
\end{align}
Now, we examine the prior term
\begin{align}
\lim_{N \to \infty} \frac{1}{N} \left. \frac{\partial \log c_{\lambda^*}(N)}{\partial n} \right|_{n=0}
:= \mathcal{G}(q_d,q) .
\end{align}
For the pattern term in \ref{eqn:real_f}, we apply the Hubbard-Stratonovich transformation \cite{10.1063/1.2917066}, the replica limit yielding
\begin{align}
\lim_{N\to\infty} \frac{1}{N} \left. \frac{\partial\log c_{\lambda^*}(P)} {\partial n} \right|_{n=0}
= \alpha_r \int\mathcal{D}t \log\Phi\left( \frac{\sqrt{q}t-\kappa} {\sqrt{q_d-q}} \right),
\end{align}
where $\mathcal{D}t = \frac{1}{\sqrt{2\pi}} \exp(-t^2/2) dt$ and $\Phi$ is the standard normal cumulative distribution function. The proof follows similarly to that in section \ref{sec:analysis}. Therefore, substituting the remaining terms back into our linear decomposition, our final quenched free energy is
\begin{align}
f(q_d,q) &= \mathcal{G}(q_d,q) + q_d\log(q_d-q)-q_d+q + \alpha_r
\int\mathcal{D}t \log\Phi\left( \frac{\sqrt{q}t-\kappa} {\sqrt{q_d q}}
\right).
\end{align}
To determine critical capacity, we differentiate in $q$ and set to zero. Moreover, we take the limit, and so
\begin{align}
\partial_q \mathcal{G}(q_d,q) + 1 - \frac{q_d}{q_d-q} + \alpha_r \frac{\partial}{\partial q} \int \mathcal{D}t \log \Phi\left( \frac{\sqrt{q} t - \kappa}{\sqrt{q_d-q}} \right) = 0.
\end{align}
Thus, 
\begin{align}
\alpha_r = \lim_{q \rightarrow 1} \frac{\frac{q}{q_d-q} - \partial_q \mathcal{G}(q_d,q)}{\displaystyle \frac{\partial}{\partial q} \int \mathcal{D}t \log \Phi\left( \frac{\sqrt{q} t - \kappa}{\sqrt{q_d-q}} \right)} 
\end{align}
as desired. In the limit, we note $q_d \rightarrow 1$ with high probability. Thus, we have the result.

\subsection{Part II: Theorem 1, complex-on-complex}
\label{app:complex}

In this section, we examine the complex-on-complex hypothesis class. We perform no restriction on pre-activation or weights. Our proof strategy here starts out similarly as before, which is classical. Our setup is quite different, so we will evaluate the quenched energy with Fourier transforms as replacement of the HCIZ step. The conclusion of the proof is like in section \ref{app:real_phase}.

\vspace{2mm}

\noindent We examine the fractional volume of the version space via Gardner volume in an unrestricted setting
\begin{align}
V_{\text{unconstrained}} = \int_{\mathbb{C}^N} \exp \left( - \Theta^{\dagger} \Theta \right) \frac{\omega^N}{N!} \prod_{\mu=1}^P H \Bigg(\text{Re}(\Psi^{\mu}) - \kappa \Bigg) ,
\end{align}
with quenched average
\begin{align}
\langle V_{\text{unconstrained}}^n \rangle = \int_{(\mathbb{C}^N)^n} \left( \bigwedge_{a=1}^n \exp \left( - \Theta_a^{\dagger} \Theta_a \right) \frac{\omega_a^N}{N!} \right) \left[ \left\langle \prod_{a=1}^n H\Big(\text{Re}(\Psi_a^\mu) - \kappa\Big) \right\rangle_Z \right]^P .
\end{align}
Applying the claim as in \ref{app:real_phase}, we get
\begin{align}
& \langle V_{\text{unconstrained}}^n \rangle  = \lim_{\epsilon \to 0^+}  \int_{(\mathbb{C}^N)^n} \left( \bigwedge_{a=1}^n \exp \left( - \Theta_a^{\dagger} \Theta_a \right) \frac{\omega_a^N}{N!} \right) \left[ \left\langle \prod_{a=1}^n H\Big(\text{Re}(\Psi_a^\mu) - \kappa\Big) \right\rangle_Z \right]^P 
\\
& \times \int_{\mathcal{H}_n} dQ \int_{i\mathcal{H}_n} d\widehat{Q}  \exp\Bigg( -N \text{Tr}\Bigg( \widehat{Q} (Q - R(\Theta)) \Bigg) \Bigg) \exp\Bigg(-\epsilon \text{Tr}(Q^2 - \widehat{Q}^2)\Bigg)
\\[2em]
= & \lim_{\epsilon \rightarrow 0^+}  \int_{\mathcal{H}_n} dQ \int_{i\mathcal{H}_n} d\widehat{Q} \int_{(\mathbb{C}^N)^n} \left( \bigwedge_{a=1}^n \exp \left( - \Theta_a^{\dagger} \Theta_a \right) \frac{\omega_a^N}{N!} \right) \left[ \left\langle \prod_{a=1}^n H\Big(\text{Re}(\Psi_a^\mu) - \kappa\Big)  \right\rangle_Z \right]^P 
\\
&  \times    \exp\Bigg( -N \text{Tr}\Big( \widehat{Q} Q \Big) \Bigg) \exp \Bigg(  N \text{Tr} \Big( \widehat{Q} R(\Theta) \Big)\Bigg) \exp\Bigg(-\epsilon \text{Tr}(Q^2 - \widehat{Q}^2)\Bigg)  .
\end{align}
The exchange is justified by Fubini's. Let us attempt to do a similar substitution as before. The line that now fails is
\begin{align}
\left[ \left\langle \prod_{a=1}^n H\Big(\text{Re}(\Psi_a^\mu) - \kappa\Big) \delta\Big(\text{Im}(\Psi_a^\mu)\Big) \right\rangle_Z \right]^P \stackrel{!}{\implies} \exp \Bigg( P G_1(Q) \Bigg) .
\end{align}
Now, we get
\begin{align}
\label{eqn:G1_complex}
\left[ \left\langle \prod_{a=1}^n H\Big(\text{Re}(\Psi_a^\mu) - \kappa\Big) \right\rangle_Z \right]^P \implies \exp \Bigg( P G_1\Big(\text{Re}(Q)\Big) \Bigg) .
\end{align}
Observe we have an asymptotic correspondence of terms for suitable $G_0, G_1$ mappings
\begin{align}
\left[ \left\langle \prod_{a=1}^n H\Big(\text{Re}(\Psi_a^\mu) - \kappa\Big) \right\rangle_Z \right]^P &\implies \exp \Bigg( P G_1(\text{Re}(Q)) \Bigg) 
\\
\int_{(\mathbb{C}^N)^n} \left( \bigwedge_{a=1}^n \exp \left( - \Theta_a^{\dagger} \Theta_a \right) \frac{\omega_a^N}{N!} \right) \exp \Bigg( N \text{Tr} \Big( \widehat{Q} R(\Theta) \Big)\Bigg) &\implies \exp \Bigg( N G_0(\widehat{Q}) \Bigg) ,
\end{align}
i.e. via definition
\begin{align}
G_0(\widehat{Q}) := \lim_{N \to \infty} \frac{1}{N} \log \left( \int_{(\mathbb{C}^N)^n} \left( \bigwedge_{a=1}^n \exp \left( - \Theta_a^{\dagger} \Theta_a \right) \frac{\omega_a^N}{N!} \right) \exp \Bigg( N \text{Tr} \Big( \widehat{Q} R(\Theta) \Big)\Bigg) \right) .
\end{align}
Substituting back in, we see
\begin{align}
\label{eqn:v_unconstrained}
\langle V_{\text{unconstrained}}^n \rangle &= \lim_{\epsilon \rightarrow 0^+} \int_{\mathcal{H}_n} dQ \int_{i\mathcal{H}_n} d\widehat{Q} \exp \Bigg( P G_1(\text{Re}(Q)) \Bigg) \exp \Bigg( N G_0(\widehat{Q}) \Bigg)
\\
& \ \ \ \ \ \ \ \ \ \times \exp\Bigg( -N \text{Tr}( Q \widehat{Q} ) \Bigg) \exp\Bigg(-\epsilon \text{Tr}(Q^2 - \widehat{Q}^2)\Bigg) .
\end{align}
This setup we have just established will provide the grounds for our proof, and this setup is nonstandard in literature and specific to our work. Thus, because of this setup, our contribution in this section in meaningful. The orthogonal invariance, as well as the use of the real operator, are specific to us.

\vspace{2mm}

\noindent While our approaches are supported via the techniques as in \ref{app:real_phase}, the region defined by the Heaviside constraints $\text{Re}(\Psi_a^\mu) > \kappa$ is not invariant under continuous orthogonal transformations $O(n)$. We will attempt to decompose our setup via Zonal polynomials, similarly to the Schur polynomials of \ref{app:real_phase}, since Zonal polynomials correspond to $O(n)$ instead of $U(n)$. In particular, the generating function $G_1(\text{Re}(Q))$ is not a class function of the orthogonal group.

\vspace{2mm}

Moreover, consider the following. Let us decompose $Q, \widehat{Q}$ into imaginary and real parts $Q = Q_R + iQ_I$, $\widehat{Q} = i\Omega_R - \Omega_I$. One of the trace terms decomposes as 
\begin{align}
\text{Tr}(Q \widehat{Q}) &= \text{Tr}\big( (Q_R + iQ_I)(i\Omega_R - \Omega_I) \big) \\
&= i\text{Tr}(Q_R \Omega_R) - \text{Tr}(Q_R \Omega_I) - \text{Tr}(Q_I \Omega_R) - i\text{Tr}(Q_I \Omega_I) \\
&= i\text{Tr}(Q_R \Omega_R) - i\text{Tr}(Q_I \Omega_I).
\end{align} 
Two terms vanish under orthogonality of symmetric and skew-symmetric matrices. Moreover, we get a decomposition
\begin{align}
\text{Tr}(Q^2 - \widehat{Q}^2) = \text{Tr}(Q_R^2 - Q_I^2 + \Omega_R^2 - \Omega_I^2). 
\end{align}
Therefore, we can immediately rewrite \ref{eqn:v_unconstrained} as
\begin{align}
& \langle V_{\text{unconstrained}}^n \rangle = \lim_{\epsilon \to 0^+}  \int_{\mathcal{S}_n} dQ_R \int_{\mathcal{S}_n} d\Omega_R \exp\Bigg(P G_1(Q_R)  - iN \text{Tr}(Q_R \Omega_R)  - \epsilon \text{Tr}(Q_R^2 + \Omega_R^2) \Bigg) \\
& \times \left[ \int_{\mathcal{A}_n} d\Omega_I \exp \Bigg( N G_0(i\Omega_R - \Omega_I)  + \epsilon \text{Tr}(\Omega_I^2) \Bigg)\int_{\mathcal{A}_n} dQ_I  \exp\Bigg( iN \text{Tr}(Q_I \Omega_I) - \epsilon \text{Tr}(-Q_I^2) \Bigg) \right].
\end{align}
where $\mathcal{S}_n$ is the real-symmetric collection and $\mathcal{A}_n$ is the real skew-symmetric. We have noted $G_1$ depends on the real part only. Using limit properties, assuming suitable hypotheses, notice
\begin{align}
\lim_{\epsilon \rightarrow 0^+} \int_{\mathcal{A}_n} dQ_I \exp\Bigg( iN \text{Tr}(\Omega_I Q_I) - \epsilon \text{Tr}(-Q_I^2) \Bigg) \rightarrow \delta(\Omega_I) 
\end{align}
(we will see this again with justification with a different integral later in the proof). We observe the above uses a trace and not a Frobenius norm, which we will see later. The skew-symmetric property is also of use in the above. Therefore, by a sifting property,
\begin{align}
\lim_{\epsilon \to 0^+} \int_{\mathcal{A}_n} d\Omega_I \exp \Bigg( N G_0(i\Omega_R - \Omega_I) + \epsilon \text{Tr}(\Omega_I^2) \Bigg) \delta(\Omega_I)  = \exp \Bigg( N G_0(i\Omega_R) \Bigg) .
\end{align}
Therefore, we can effectively bypass the imaginary portion in these terms, and thus we may consider a Weyl integration change of variables with respect to $O(n)$ instead of $U(n)$. This permits the Zonal polynomial expansion.

\vspace{2mm}

Therefore, we examine the integrals again under a Weyl integration change of variables via the change in volume element $dQ = |\Delta(\Lambda)| \prod_{i=1}^n d\lambda_i  dO$
\begin{align}
& \langle V_{\text{unconstrained}}^n \rangle  \propto \lim_{\epsilon \rightarrow 0^+} \int_{\mathbb{R}^n} d\Lambda \int_{(i\mathbb{R})^n} d\widehat{\Lambda} |\Delta(\Lambda)| |\Delta(\widehat{\Lambda})| 
\\
& \ \ \ \ \ \ \ \ \ \times \exp \Bigg( N G_0(\widehat{\Lambda}) \Bigg) \exp\Bigg(-\epsilon \text{Tr}(\Lambda^2 - \widehat{\Lambda}^2)\Bigg)
\\
& \ \ \ \ \ \ \ \ \ \times \vast[ \int_{O(n)} d\widehat{O} \int_{O(n)} dO \exp \Bigg( -N \text{Tr}\Big(\Lambda (O^T \widehat{O}) \widehat{\Lambda} (O^T \widehat{O})^T \Big) \Bigg) \exp \Bigg( P G_1(O \Lambda O^T) \Bigg) \vast]
\\[2em]
& = \lim_{\epsilon \rightarrow 0^+} \int_{\mathbb{R}^n} d\Lambda \int_{(i\mathbb{R})^n} d\widehat{\Lambda} |\Delta(\Lambda)| |\Delta(\widehat{\Lambda})| 
\\
& \ \ \ \ \ \ \ \ \ \times \exp \Bigg( N G_0(\widehat{\Lambda}) \Bigg) \exp\Bigg(-\epsilon \text{Tr}(\Lambda^2 - \widehat{\Lambda}^2)\Bigg)
\\
& \ \ \ \ \ \ \ \ \ \times \vast[ \int_{O(n)} dR \exp \Bigg( -N \text{Tr}\Big(\Lambda R \widehat{\Lambda} R^T \Big) \Bigg) \vast] \vast[ \int_{O(n)} dO \exp \Bigg( P G_1(O \Lambda O^T) \Bigg) \vast] .
\end{align}

\vspace{2mm}

\noindent To permit a polynomial expansion, we define
\begin{align}
\label{eqn:O_invariance}
\exp \Bigg( P \widetilde{G}_1(\Lambda) \Bigg) := \int_{O(n)} dO \exp \Bigg( P G_1(O \Lambda O^T) \Bigg) .
\end{align}
Also, $G_0(\widehat{Q})$ is orthogonally invariant by construction ($G_0(O \widehat{Q} O^T) = G_0(\widehat{Q})$). We forced $G_1$ to be invariant via \ref{eqn:O_invariance}.

\vspace{2mm}

\vspace{2mm}

\noindent Because $\widetilde{G}_1(\Lambda)$ and $G_0(\widehat{\Lambda})$ are now both orthogonally invariant functions of symmetric matrices, they can be expanded formally into a basis of orthogonally invariant Zonal polynomials \cite{736f114d-4335-3bf4-b381-39abefbb8981} \cite{Muirhead1982AspectsOM}. Let us revert back to our original domains of integration but solely decompose the $G_0, G_1$ terms. Thus, under a valid Zonal polynomial decomposition, the integral becomes
\begin{align}
\label{eqn:zonal_decomposition}
\langle V_{\text{unconstrained}}^n \rangle &= \lim_{\epsilon \rightarrow 0^+} \int_{\mathcal{H}_n} dQ \int_{i\mathcal{H}_n} d\widehat{Q} \sum_{\lambda} \sum_{\mu} c_{\lambda}(P) c_{\mu}(N) Z_{\lambda}(\text{Re}(Q)) Z_{\mu}(\text{Re}(\widehat{Q}))
\\
& \ \ \ \ \ \ \ \ \ \times \exp\Bigg( -N \text{Tr}( Q \widehat{Q} ) \Bigg) \exp\Bigg(-\epsilon \text{Tr}(Q^2 - \widehat{Q}^2)\Bigg) .
\end{align}

\vspace{2mm}

Perform variable reassignment $ \widehat{Q} = i \Omega $, thus $ d\widehat{Q} = i^{n^2} d\Omega $ and $ Z_\mu(\widehat{Q}) = Z_\mu(i\Omega) = i^{|\mu|} Z_\mu(\Omega) $. Pulling the sums out of the integral and the limit, justifying similarly as in Appendix \ref{app:real_phase}
\begin{align}
\label{eqn:v_unconstrained_zonal}
& \langle V_{\text{unconstrained}}^n \rangle = \lim_{\epsilon \rightarrow 0^+} \sum_{\lambda} \sum_{\mu} c_{\lambda}(P) c_{\mu}(N) i^{n^2 + |\mu|} \int_{\mathcal{H}_n} dQ \int_{\mathcal{H}_n} d\Omega Z_{\lambda}(\text{Re}(Q)) Z_{\mu}(\text{Re}(\Omega))\\
& \ \ \ \ \ \ \ \ \ \ \ \ \ \ \ \ \ \ \ \ \ \ \ \ \ \ \ \ \ \ \ \ \ \ \ \ \ \ \ \ \ \ \ \ \  \times \exp\Bigg( -i N \text{Tr}( Q \Omega ) \Bigg) \exp\Bigg(-\epsilon \text{Tr}(Q^2 + \Omega^2)\Bigg)
\\
& =  \sum_{\lambda} \sum_{\mu} c_{\lambda}(P) c_{\mu}(N) i^{n^2 + |\mu|}  \lim_{\epsilon \rightarrow 0^+} \int_{\mathcal{H}_n} dQ Z_{\lambda}(\text{Re}(Q)) \int_{\mathcal{H}_n} d\Omega  Z_{\mu}(\text{Re}(\Omega)) 
\\
& \ \ \ \ \ \ \ \ \ \ \ \ \ \ \ \ \ \ \ \ \ \ \ \ \ \ \ \ \ \ \ \ \ \ \ \ \ \ \ \ \ \ \ \ \ \exp \Bigg( - \epsilon \text{Tr}(\Omega^2) \Bigg) \exp \Bigg(-i N \text{Tr}(Q\Omega)\Bigg)  \exp\Bigg( -\epsilon \text{Tr}(Q^2) \Bigg) .
\end{align}
Recall the Fourier transform definition
\begin{align}
\mathcal{F}[f](K) = \int_{\mathcal{H}_n}d\Omega   f(\Omega) \exp\Bigg(-i \text{Tr}(K \Omega)\Bigg) 
\end{align}
under a trace inner product. Therefore, we can note
\begin{align}
\mathcal{F} \Bigg[ Z_{\mu}(\text{Re}(\Omega)) \exp\Bigg(-\epsilon \text{Tr}(\Omega^2)\Bigg) \Bigg] (NQ) = \int_{\mathcal{H}_n}d\Omega Z_{\mu}(\text{Re}(\Omega)) \exp\Bigg(-\epsilon \text{Tr}(\Omega^2)\Bigg) \exp\Bigg(-i N \text{Tr}(Q\Omega)\Bigg)  .
\end{align}
The Gaussian regularizer $\exp(-\epsilon \text{Tr}(\Omega^2))$ is a function in Schwartz space $\mathcal{S}$ that converges to the constant $1$ in the topology of $\mathcal{S}'$. In distribution theory, we have the rule when $N$ is large
\begin{align}
\mathcal{F}\Bigg[ P(\Omega) g(\Omega) \Bigg](NQ) = P\left(\frac{i}{N} \nabla_Q\right) \mathcal{F}\Bigg[ g(\Omega) \Bigg](NQ) 
\end{align}
for polynomial $P$ and $P(D)$ is the polynomial version of the differential operator. Therefore, we have
\begin{align}
& \int_{\mathcal{H}_n}d\Omega Z_{\mu}(\text{Re}(\Omega)) \exp\Bigg(-\epsilon \text{Tr}(\Omega^2)\Bigg) \exp\Bigg(-i N \text{Tr}(Q\Omega)\Bigg) 
\\
& = Z_{\mu}\left(\frac{i}{N} \nabla_Q\right) \Bigg[ \int_{\mathcal{H}_n} d\Omega \exp\Bigg(-\epsilon \text{Tr}(\Omega^2)\Bigg) \exp\Bigg(-i N \text{Tr}(Q\Omega)\Bigg)  \Bigg] .
\end{align}
Thus, taking the limit
\begin{align}
\lim_{\epsilon \to 0^+} \Bigg[ \int_{\mathcal{H}_n} d\Omega \exp\Bigg(-\epsilon \text{Tr}(\Omega^2)\Bigg) \exp\Bigg(-i N \text{Tr}(Q\Omega)\Bigg)  \Bigg] = \left(\frac{2\pi}{N}\right)^{n^2} \delta(Q) .
\end{align}
The above uses the fact that
\begin{align}
 I_\epsilon(Q) = \int_{\mathbb{R}^{n^2}} d\Omega \exp\Big(-\epsilon ||\Omega||^2 - i N \langle Q, \Omega \rangle \Big)  
\end{align}
which evaluates to the Gaussian 
\begin{align}
 I_\epsilon(Q) = \left( \frac{\pi}{\epsilon} \right)^{\frac{n^2}{2}} \exp\left( - \frac{N^2 ||Q||^2}{4\epsilon} \right)  .
\end{align}
Therefore, we examine
\begin{align}
\lim_{\epsilon \to 0^+} I_\epsilon(Q) &= \lim_{\epsilon \to 0^+} \left( \frac{\pi}{\epsilon} \right)^{\frac{n^2}{2}} \exp\left( - \frac{N^2 ||Q||^2}{4\epsilon} \right)
\\
&= \left[ \int_{\mathbb{R}^{n^2}} dQ \left( \frac{\pi}{\epsilon} \right)^{\frac{n^2}{2}} \exp\left( - \frac{N^2 ||Q||^2}{4\epsilon} \right)  \right] \delta(Q)
\\
&= \left[ \left( \frac{\pi}{\epsilon} \right)^{\frac{n^2}{2}} \left( \frac{\sqrt{4\epsilon}}{N} \right)^{n^2} \int_{\mathbb{R}^{n^2}} dU \exp(-||U||^2)  \right] \delta(Q)
\\
&= \left[ \left( \frac{\pi}{\epsilon} \right)^{\frac{n^2}{2}} \left( \frac{4\epsilon}{N^2} \right)^{\frac{n^2}{2}} \pi^{\frac{n^2}{2}} \right] \delta(Q)
\\
&= \left( \frac{2\pi}{N} \right)^{n^2} \delta(Q) .
\end{align}
We have used the rule in distribution theory $ \lim_{\epsilon \to 0} f_\epsilon(x) = \left[ \lim_{\epsilon \rightarrow 0} \int_{\mathbb{R}^d} dx' f_\epsilon(x')  \right] \delta(x) $ with concentration of mass (see \cite{stefani2025weightedversionbbmformula} \cite{gennaioli2025sharpconditionsbbmformula} for relevant literature). This implies
\begin{align}
\lim_{\epsilon \to 0^+} \int_{\mathcal{H}_n} d\Omega  Z_{\mu}(\text{Re}(\Omega)) \exp\Bigg(-\epsilon \text{Tr}(\Omega^2)\Bigg) \exp\Bigg(-i N \text{Tr}(Q\Omega)\Bigg)  = \left(\frac{2\pi}{N}\right)^{n^2} Z_{\mu}\left(\frac{i}{N} \nabla_Q\right)  \delta(Q) .
\end{align}
Returning to the original set of integrals and evaluating,
\begin{align}
& \lim_{\epsilon \to 0^+} \int_{\mathcal{H}_n}  dQ \left[ \left(\frac{2\pi}{N}\right)^{n^2} Z_{\mu}\left( \frac{i}{N} \nabla_Q \right) \delta(Q) \right] Z_{\lambda}(\text{Re}(Q)) \exp  \Bigg( -\epsilon \text{Tr}(Q^2) \Bigg) 
\\
& \stackrel{(1)}{=} \left(\frac{2\pi}{N}\right)^{n^2} \lim_{\epsilon \to 0^+} \int_{\mathcal{H}_n} dQ \delta(Q) Z_{\mu}\left( -\frac{i}{N} \nabla_Q \right) \Bigg[ Z_{\lambda}(\text{Re}(Q)) \exp \Bigg( -\epsilon \text{Tr}(Q^2) \Bigg) \Bigg]
\\
& \stackrel{(2)}{=} \left(\frac{2\pi}{N}\right)^{n^2} \left(-\frac{i}{N}\right)^{|\mu|} \lim_{\epsilon \to 0^+} \Bigg[ Z_{\mu}( \nabla_Q ) \Bigg( Z_{\lambda}(\text{Re}(Q)) \exp \Bigg( -\epsilon \text{Tr}(\text{Re}(Q)^2 + \text{Im}(Q)^2) \Bigg) \Bigg) \Bigg] \Bigg|_{Q=0}
\\
& \stackrel{\text{evaluate}}{=} \left(\frac{2\pi}{N}\right)^{n^2} \left(-\frac{i}{N}\right)^{|\mu|} \Bigg[ Z_{\mu}( \nabla_{\text{Re}(Q)} ) Z_{\lambda}(\text{Re}(Q)) \Bigg] \Bigg|_{\text{Re}(Q)=0} 
\\
\label{eqn:change_basis}
& \stackrel{\text{change of basis}}{=} \left(\frac{2\pi}{N}\right)^{n^2} \left(-\frac{i}{N}\right)^{|\mu|} \sum_{\nu} \theta_{\mu}^\nu \Bigg[ Z_{\nu}( \nabla_{\text{Re}(Q)} ) Z_{\lambda}(\text{Re}(Q)) \Bigg] \Bigg|_{\text{Re}(Q)=0} 
\\
& \stackrel{\text{orthogonality}}{=} \left(\frac{2\pi}{N}\right)^{n^2} \left(-\frac{i}{N}\right)^{|\mu|} \sum_{\nu} \theta_{\mu}^\nu \delta_{\lambda,\nu} z_\lambda
\\
&= \left(\frac{2\pi}{N}\right)^{n^2} \left(-\frac{i}{N}\right)^{|\mu|} \theta_{\mu}^\lambda z_\lambda .
\end{align}
(1) follows by integration by parts, and (2) follows by scaling and integrating, using the sifting property $\int V(Q)\delta(Q) dQ = f(0)$ \cite{WeissteinSifting}. The constant coming out in (2) follows from the homogeneity property of polynomials. Therefore, we may return to our original integrals again and substitute in what we found. Here, $\theta$ is a change-of-basis translation matrix. Thus, we see after returning to \ref{eqn:v_unconstrained_zonal}, using the result of $z_{\lambda}$,
\begin{align}
\label{eqn:vn_unconstrained_changeofbasis}
& \langle V_{\text{unconstrained}}^n \rangle  = \sum_{\lambda} \sum_{\mu} c_{\lambda}(P) c_{\mu}(N) i^{n^2 + |\mu|} \lim_{\epsilon \rightarrow 0^+} \int_{\mathcal{H}_n} dQ Z_{\lambda}(\text{Re}(Q)) \int_{\mathcal{H}_n} d\Omega Z_{\mu}(\text{Re}(\Omega)) 
\\
& \ \ \ \ \ \ \ \ \ \ \ \ \ \ \ \ \ \ \ \ \ \ \times \exp \Bigg( - \epsilon \text{Tr}(\Omega^2) \Bigg) \exp \Bigg(-i N \text{Tr}(Q\Omega) \Bigg) \exp\Bigg( -\epsilon \text{Tr}(Q^2) \Bigg)
\\[1em]
&= \sum_{\lambda} \sum_{\mu} c_{\lambda}(P) c_{\mu}(N) \left(\frac{2\pi}{N}\right)^{n^2} \left[ i^{|\mu|} \left(-\frac{i}{N}\right)^{|\mu|} \right] \theta_{\mu}^{\lambda} \prod_{s \in \lambda} \Big( 2a(s) + l(s) + 1 \Big) \Big( 2a(s) + l(s) + 2 \Big)
\\[1em]
&= \underbrace{ \left(\frac{2\pi}{N}\right)^{n^2} }_{\mathcal{Z}_{N,n}} \sum_{\lambda} \sum_{\mu} c_{\lambda}(P) c_{\mu}(N) \frac{\theta_{\mu}^{\lambda}}{N^{|\mu|}} \prod_{s \in \lambda} \Big( 2a(s) + l(s) + 1 \Big) \Big( 2a(s) + l(s) + 2 \Big)  .
\end{align}
For simplicity, omit the $i^{n^2}$, which vanishes in the thermodynamic limit. We can also note 
\begin{align}
i^{|\mu|} \left(-\frac{i}{N}\right)^{|\mu|} = \left( i \cdot -\frac{i}{N} \right)^{|\mu|} = \left( - \frac{i^2}{N} \right)^{|\mu|} .
\end{align}
Since $i^2 = -1$, the numerator becomes $-(-1) = 1$, leaving
\begin{align}
\left( \frac{1}{N} \right)^{|\mu|} = \frac{1}{N^{|\mu|}} .
\end{align}
Note that the Kronecker delta $\delta_{\lambda,\nu}$ enforces $\lambda = \nu$. Recall the identity $x = \exp \log x$. Thus, we exponentiate and splitting the log using log properties, we see from \ref{eqn:vn_unconstrained_changeofbasis}
\begin{align}
& \langle V_{\text{unconstrained}}^n \rangle = \sum_{\lambda} \sum_{\mu} \exp \left\{ \log \left( \mathcal{Z}_{N,n} c_{\lambda}(P) c_{\mu}(N) \theta_{\mu}^{\lambda} \frac{1}{N^{|\mu|}} \prod_{s \in \lambda} \left( 2a(s) + l(s) + 1 \right) \left( 2a(s) + l(s) + 2 \right) \right) \right\} \\
&= \sum_{\lambda} \sum_{\mu} \exp \left\{ \log \mathcal{Z}_{N,n} + \log c_{\lambda}(P) + \log c_{\mu}(N) + \log \theta_{\mu}^{\lambda} +  \sum_{s \in \lambda} \log \Big( (2a(s) + l(s) + 1)(2a(s) + l(s) + 2) \Big)  - |\mu| \log N  \right\}
\\
& := \sum_{\lambda} \sum_{\mu} \exp \left \{ S_{\text{unconstrained}}(\lambda, \mu, n) \right\}.
\end{align}
By a discrete version of Laplace's method,
\begin{align}
\label{eqn:laplace_complex}
& \sum_{\lambda} \sum_{\mu} \exp \left\{ S_{\text{unconstrained}}(\lambda, \mu, n) \right\}
\\
& \xrightarrow{N \to \infty}
\left( \frac{1}{\Delta x} \right)^n \underbrace{ \sqrt{\frac{(2\pi)^n}{\det \Big( - \mathcal{H}_{\lambda^*, \mu^*}(S_{\text{unconstrained}}) \Big)}} }_{= C(n)}
\exp \left\{ \mathop{\text{extr}}_{\lambda^*, \mu^*} S_{\text{unconstrained}}(\lambda^*, \mu^*, n) \right\},
\end{align}
where the saddle point partitions $(\lambda^*, \mu^*)$ satisfy the extremal conditions. Note that we are not technically evaluating the limit yet, but rather it is an asymptotic expansion. Evaluating the thermodynamic limit,
\begin{align}
\label{eqn:f_complex}
f = -\lim_{(N, P, \alpha) \to \infty} \frac{1}{N} \lim_{n \to 0} \frac{\log \mathcal{Z}^n}{n}
= -\lim_{(N, P, \alpha) \to \infty} \frac{1}{N}
\left(
  \left. \frac{\partial S_{\text{unconstrained}}(\lambda^*, \mu^*, n)}{\partial n} \right|_{n=0}
+ \left. \frac{\partial \log C(n)}{\partial n} \right|_{n=0}
\right).
\end{align}
This follows from the formal definition of the partial derivative, and $S_{\text{unconstrained}}(\lambda^*, \mu^*, 0) = 0$. By linearity of the derivative, we expand \ref{eqn:f_complex}
\begin{align}
&\underbrace{
  \left. \frac{\partial \log \mathcal{Z}_{N,n}}{\partial n} \right|_{n=0}
}_{\mathcal{S}_{\text{constant}}}
+
\underbrace{
  \left. \frac{\partial \log c_{\mu^*}(N)}{\partial n} \right|_{n=0}
}_{\mathcal{S}_{\text{prior}}}
+
\underbrace{
  \left. \frac{\partial \log \theta_{\mu^*}^{\lambda^*}}{\partial n} \right|_{n=0}
}_{\mathcal{S}_{\text{branch}}}
\\
&+
\underbrace{
  \left. \frac{\partial}{\partial n} \sum_{s \in \lambda^*} \log \big( (2a(s) + l(s) + 1)(2a(s) + l(s) + 2) \big) \right|_{n=0}
}_{\mathcal{S}_{\text{mixing}}}
-
\underbrace{
  \left. \frac{\partial}{\partial n} \left( |\mu^*| \log N \right) \right|_{n=0}
}_{\mathcal{S}_{\text{scaling}}}
\\
&+
\underbrace{
  \left. \frac{\partial \log c_{\lambda^*}(P)}{\partial n} \right|_{n=0}
}_{\mathcal{S}_{\text{patterns}}(\alpha_c)}
+
\underbrace{
  \left. \frac{\partial \log C(n)}{\partial n} \right|_{n=0}
}_{\mathcal{S}_{\text{fluctuations}}}
= f.
\end{align}
For a replica-symmetric rectangular partition evaluated at $n=0$, the mixing term evaluates to $\log((2l^*)!)$ and the scaling term evaluates to $l^* \log N$. By Stirling's approximation, we get
\begin{align}
\log((2l^*)!) - l^* \log N \approx 2l^* \log(2l^*) - 2l^* - l^* \log N.
\end{align}
In the thermodynamic limit, the entropy of the branching coefficient \cite{Parisi_2015} \cite{kabluchko2014generalizedrandomenergymodel} is given in Appendix \ref{app:proofs}. Because they scale slower than $\mathcal{O}(N)$, we will omit $\mathcal{S}_{\text{constant}}$ and $\mathcal{S}_{\text{fluctuations}}$ in the final computation as a rough estimate since they vanish in the thermodynamic limit and after the final differentiation. We apply the $1/N$ scaling and set $l^* = \rho N$.

\vspace{2mm}

Note that $l^* = \rho N$ enforces the dominant partition, referring to $\lambda$, to scale macroscopically with the size of the network $N$. This equation corresponds to the replica symmetric ansatz. Because we justified 
\begin{align}
q_d = \frac{1}{N} \| \Theta \|_2^2 \xrightarrow{N \to \infty} 1 \quad \text{almost surely} ,
\end{align}
and using similarly strategies to the finale of the proof of Appendix \ref{app:real_phase}, we will allow $l^* = \rho N$ to hold. This will let us calculate our final limit. Note the partition macroscopic scaling $l^* = \rho N$ is more relevant to random matrix theory rather than Gardner volume calculation, and is suitable for us. We refer to \cite{GUIONNET2005435} \cite{Gardner_1988} for some relevant literature.

\vspace{2mm}

\noindent Here we use Lemma 1, Lemma 2, and Lemma 3. Taking the limit, and noting $ \mathcal{S}_{\text{patterns}}(\alpha_c) = -\rho N \log N + \alpha_c N \mathcal{G}_c(\rho,\kappa) - N v +  o(N) $ (the leading order term and the $v$ control terms come from a volume term and $\mathcal{G}_c(\rho,\kappa)$ comes from a $\mathbb{X}$ indicator or Gibbs weight term (see Lemma 2 in \ref{app:proofs}; discussed variously in \cite{James1964DistributionsOM} \cite{wang2024explicitformulazonalpolynomials} \cite{malatesta2025highdimensionalmanifoldsolutionsneural} \cite{baldassi2023typicalatypicalsolutionsnonconvex}))
\begin{align}
&\lim_{N \to \infty} \frac{1}{N} \Bigg[ \underbrace{
  \Big( - \rho N \log(\rho N) + \rho N \log N + \mathcal{B}(\rho N) \Big) }_{\mathcal{S}_{\text{branch}}}
\\
& \qquad\qquad \qquad\qquad +
  \underbrace{
    \Big( 2\rho N \log(2\rho N) - 2\rho N - \rho N \log N  + N[(1+\rho)\log(1+\rho) - \rho\log\rho] \Big)
  }_{\mathcal{S}_{\text{mixing}} + \mathcal{S}_{\text{scaling}} + \mathcal{S}_{\text{prior}}}
\\
& \qquad\qquad \qquad\qquad \qquad\qquad \qquad\qquad+
  \underbrace{
    \Big( - \rho N \log N - \log V(Q) + \alpha_c N \mathcal{G}_c(\rho,\kappa) - N v + o(N) \Big)
  }_{\mathcal{S}_{\text{patterns}}}
\Bigg]
\\
= \ & \lim_{N \to \infty} \frac{1}{N} \Bigg[
  - \rho N \log \rho - \rho N \log N + \rho N \log N + \mathcal{B}(\rho N)
  + 2\rho N \log 2 + N[(1+\rho)\log(1+\rho) - \rho\log\rho]
\\
&\qquad\qquad
  + 2\rho N \log \rho + 2\rho N \log N - 2\rho N - \rho N \log N - \rho N \log N + \alpha_c N \mathcal{G}_c(\rho,\kappa) - N v + o(N)
\Bigg].
\end{align}
Note that we have relaxed a sign convention on $v$, which is unimportant. Grouping the terms, we observe
\begin{align}
&= \lim_{N \to \infty} \frac{1}{N} \Bigg[
  \rho N \log \rho + 2\rho N \log 2 - 2\rho N + \mathcal{B}(\rho N) + \alpha_c N \mathcal{G}_c(\rho,\kappa) - N v + N[(1+\rho)\log(1+\rho) - \rho\log\rho]+ o(N)
\Bigg]
\\
&= \rho \log \rho + 2\rho \log 2 - 2\rho + \mathcal{B}(\rho) + \alpha_c {\mathcal{G}_c} - v + (1+\rho)\log(1+\rho) - \rho\log\rho .
\end{align}
As before, we must differentiate in $\rho$ and set to zero. We get
\begin{align}
\log \rho + 2 \log 2 - 1 + \mathcal{B}'(\rho) + \alpha_c \partial_{\rho} \mathcal{G}_c(\rho,\kappa) + \log \left( \frac{1+\rho}{\rho} \right) = 0 .
\end{align}
Rearranging, critical capacity is
\begin{align}
\alpha_c = \lim_{\rho \rightarrow \rho_c} \frac{1  - 2 \log 2 + \log ( 1 + \rho )  - \mathcal{B}'(\rho)}{\frac{\partial}{\partial \rho} \mathcal{G}_c(\rho,\kappa)} ,
\end{align}
and this concludes the proof.

\noindent $ \square$

\section{Additional proofs}
\label{app:proofs}

\subsection{Complex-on-complex proofs}

In this section, we prove Lemmas 1-3. Our proofs are mostly formal, although slightly more informal than the main proofs in Appendices \ref{app:real_phase}, \ref{app:complex}.

\vspace{2mm}

\textbf{Lemma 1.} Let $\mathcal{S}_{\text{branch}}$ be as in \ref{app:complex}. It follows that
\begin{align}
\mathcal{S}_{\text{branch}} = -l^* \log(l^*) + l^* \log(N) + \mathcal{B}(l^*) .
\end{align}

\textit{Proof.} Let us begin with the definition of change-of-basis $\theta$ that we defined in section \ref{app:real_phase}. $\theta$ measures a way to translate between two volumes of the neural network's parameter space. We are moving from a standard coordinate basis to representations via the orthogonal group $O(n)$, which is what the Zonal polynomials represent. We have the branching coefficient obeys, which is a change of basis term, under assumption with high probability
\begin{align}
\label{eqn:theta_form}
\theta_{\mu^*}^{\lambda^*}(n) \approx \underbrace{ \left( \frac{N^{l^*}}{l^*!} \right)^n }_{\text{assumed combinatorial scaling}} \underbrace{ e^{n \widetilde{\mathcal{B}}(l^*)} }_{\text{extensivity assumption}}  ,
\end{align}
where $\widetilde{\mathcal{B}}(l^*)$ is a suitable function. We keep $\widetilde{\mathcal{B}}$ highly generalized, so the above \ref{eqn:theta_form} is an assumption.

\vspace{2mm}

Now, we take the logarithm of \ref{eqn:theta_form}
\begin{align}
\log \theta_{\mu^*}^{\lambda^*}(n) = n l^* \log N - n \log(l^*!) + n \widetilde{\mathcal{B}}(l^*)  .
\end{align}
Next, take the partial derivative,
\begin{align}
\frac{\partial \log \theta_{\mu^*}^{\lambda^*}}{\partial n} = l^* \log N - \log(l^*!) + \widetilde{\mathcal{B}}(l^*)  .
\end{align}
Thus
\begin{align}
\label{eqn:s_branch_closed_form}
\mathcal{S}_{\text{branch}} = l^* \log N - \log(l^*!) + \widetilde{\mathcal{B}}(l^*)  .
\end{align}
We can expand the factorial using Stirling's approximation, $\log(x!) \approx x \log x - x$, $\log(l^*!) \approx l^* \log(l^*) - l^*$. Substituting this back into our equation \ref{eqn:s_branch_closed_form}
\begin{align}
\label{eqn:s_branch_final}
\mathcal{S}_{\text{branch}} & = l^* \log N - \Big( l^* \log(l^*) - l^* \Big) + \widetilde{\mathcal{B}}(l^*)  
\\
& = -l^* \log(l^*) + l^* \log N + \Big( l^* + \widetilde{\mathcal{B}}(l^*) \Big)  
\\
& = -l^* \log(l^*) + l^* \log(N) + \mathcal{B}(l^*)  ,
\end{align}
where $\mathcal{B}(l^*) = l^* + \widetilde{\mathcal{B}}(l^*)$ is defined appropriately.

\noindent $\square $

\vspace{2mm}

\textbf{Lemma 2.} Define $\mathcal{S}_{\text{patterns}}$ as in \ref{app:complex}. It follows that
\begin{align}
\mathcal{S}_{\text{patterns}}(\alpha_c) = -\rho N \log N + N v(Q^*) + \alpha_c N \mathcal{G}_c +  o(N) .    
\end{align}

\vspace{2mm}

\textit{Proof.} The Zonal polynomial coefficients closed form come from \ref{eqn:zonal_decomposition}, which can be written (see \cite{James1964DistributionsOM} \cite{wang2024explicitformulazonalpolynomials} for relevant literature)
\begin{align}
\label{eqn:c_closed_form}
c_{\lambda^*}(P) = \mathbb{E}_{\xi} \left[ \int d\mu_{\lambda^*}(\Theta^1, \dots, \Theta^n) \prod_{\mu=1}^P \prod_{a=1}^n \mathbb{X}(\Theta^a, \xi^\mu) \right].
\end{align}
Here, $\mathbb{X}$ is an indicator or Gibbs weight prototypical as in spin glass literature as in \cite{malatesta2025highdimensionalmanifoldsolutionsneural} \cite{baldassi2023typicalatypicalsolutionsnonconvex}. It is nontrivial to see why the above is true so we elaborate on this. Working backwards on our Zonal polynomials, we recall via definition of \ref{eqn:G1}
\begin{align}
\exp \Bigg( P G_1(\text{Re}(Q)) \Bigg) \impliedby \left[ \left\langle \prod_{a=1}^n H\Big(\text{Re}(\Psi_a^\mu) - \kappa\Big) \right\rangle_Z \right]^P .
\end{align}
\begin{enumerate}
\item[$\text{i}.$] $\mathbb{X}(\Theta^a, \xi^\mu)$ is our Heaviside step function $H(\text{Re}(\Psi_a^\mu) - \kappa)$.
\item[$\text{ii}.$] The full version space is $\prod \prod \mathbb{X}$, enforcing the constraints across all $n$ replicas for $P$ patterns.     \item[$\text{iii}.$] Moreover, $\mathbb{E}_\xi$ is simply the quenched average. \end{enumerate}
We introduce an overlap matrix $Q_{ab} = \frac{1}{N} (\Theta^a)^\dagger \Theta^b$. We can note, similarly to our claim as in \ref{app:real_phase},
\begin{align}
1 & = \int \mathcal{D}Q \delta_{\mathcal{H}_n} \left(Q - \frac{1}{N}\Theta^\dagger \Theta\right) 
\\
&  := \int \mathcal{D}Q \prod_{a=1}^n \delta\left(Q_{aa} - \frac{1}{N} (\Theta^a)^\dagger \Theta^a\right) \prod_{a < b} \delta_{\mathbb{C}}\left(Q_{ab} - \frac{1}{N} (\Theta^a)^\dagger \Theta^b\right) .
\end{align}
Inserting 1 in \ref{eqn:c_closed_form} like in our claim,
\begin{align}
c_{\lambda^*}(P) &= \mathbb{E}_{\xi} \left[ \int d\mu_{\lambda^*}(\Theta) \int \mathcal{D}Q \delta_{\mathcal{H}_n} \left(Q - \frac{1}{N}\Theta^\dagger \Theta\right)  \prod_{\mu=1}^P \prod_{a=1}^n \mathbb{X}(\Theta^a, \xi^\mu) \right].
\end{align}
Thus, applying commutativity and standard probability laws,
\begin{align}
\label{eqn:c_manipulated}
c_{\lambda^*}(P) &= \int \mathcal{D}Q \underbrace{ \int d\mu_{\lambda^*}(\Theta) \delta_{\mathcal{H}_n} \left(Q - \frac{1}{N}\Theta^\dagger \Theta\right)  }_{=\text{volume form}} \left( \mathbb{E}_{\xi} \left[ \prod_{a=1}^n \mathbb{X}(\Theta^a, \xi) \right] \right)^P.
\end{align}
Because of the argument in Appendix \ref{app:real_phase}, the dependence of the expectation on $\Theta$ is fully parameterized by $Q$ (see Appendix \ref{app:discussion}). We can thus define $\mathcal{G}_c$ evaluated on this constraint, so because of the Dirac delta,
\begin{align}
\mathcal{G}_c(Q) := \frac{1}{n} \log \mathbb{E}_{\xi} \left[ \prod_{a=1}^n \mathbb{X}(\Theta^a, \xi) \right]_{ \frac{1}{N}\Theta^\dagger \Theta = Q} .
\end{align}
noting existence under the convention $\log(0) = -\infty$. This allows us to substitute the expectation with $e^{n \mathcal{G}_c(Q)}$ and factor it out of the $d\mu_{\lambda^*}(\Theta)$ integral. Let us define the volume term
\begin{align}
M_{\lambda^*}(Q, n) := \int d\mu_{\lambda^*}(\Theta)\delta_{\mathcal{H}_n} \left(Q - \frac{1}{N}\Theta^\dagger \Theta\right) .
\end{align}
Thus, we can rewrite $c_{\lambda^*}(P)$ as in \ref{eqn:c_manipulated} as
\begin{align}
c_{\lambda^*}(P) = \int \mathcal{D}Q  M_{\lambda^*}(Q, n)  e^{n P \mathcal{G}_c(Q)} = \int \mathcal{D}Q \exp\Big( \log M_{\lambda^*}(Q, n) + n \alpha_c N \mathcal{G}_c(Q) \Big).
\end{align}
In the thermodynamic limit, we get the saddle point approximation via Laplace's method 
\begin{align}
c_{\lambda^*}(P) \propto \exp\Big( \log M_{\lambda^*}(Q^*, n) + n \alpha_c N \mathcal{G}_c(Q^*) \Big).
\end{align}
We have omitted constants, which are annihilated in the thermodynamic limit as before. The first term evaluates asymptotically to under the log-exponential trick for some $f(Q,n)$
\begin{align}
\label{eqn:volume_M}
M_{\lambda^*}(Q^*, n) \approx  N^{-n l^*} f(Q^*,n).
\end{align}
Substituting $l^* = \rho N$, we have
\begin{align}
\log M_{\lambda^*}(Q^*, n) = -n \rho N \log N + \log f(Q^*,n).
\end{align}
Therefore, the action evaluated at the saddle point is
\begin{align}
\mathcal{S}(n, \alpha_c) = -n \rho N \log N + \log f(Q^*,n) + n \alpha_c N \mathcal{G}_c(Q^*).
\end{align}
Taking the replica limit gives
\begin{align}
\lim_{n \to 0} \frac{\partial}{\partial n} \Big( -n \rho N \log N + \log f(Q^*,n) + n \alpha_c N \mathcal{G}_c(Q^*) \Big) = -\rho N \log N + N v(Q^*) + \alpha_c N \mathcal{G}_c(Q^*).
\end{align}
$N v(Q^*)$ is a derivative that corresponds to the $\log f$ term, since $\log f$ is macroscopic, thus $N$ appears.

\noindent $\square$

\vspace{2mm}

\textit{Remark.} Equation \ref{eqn:volume_M} comes from the fact that integrating the Zonal polynomial $Z_{\lambda^*}(\Theta^\dagger \Theta)$ of degree $l^*$ over the unconstrained complex measure produces a characteristic ratio of Gamma functions
\begin{align}
\int d\mu_0 Z_{\lambda^*}(\Theta^\dagger \Theta) \propto \frac{\Gamma(N)}{\Gamma(N + l^*)} .
\end{align}
In the thermodynamic limit, we apply Stirling's approximation ($\Gamma(z) \sim z^z e^{-z}$)
\begin{align}
\label{eqn:gamma_frac}
\frac{\Gamma(N)}{\Gamma(N + l^*)} \approx \frac{N^N e^{-N}}{(N + l^*)^{N + l^*} e^{-(N + l^*)}} .
\end{align}
We can pull out $N$ from the denominator
\begin{align}
(N + l^*)^{N + l^*} = N^{N + l^*} \left(1 + \frac{l^*}{N}\right)^{N + l^*} .
\end{align}
Substituting this back into our ratio,
\begin{align}
\frac{\Gamma(N)}{\Gamma(N + l^*)} &\approx \frac{N^N e^{-N}}{N^{N + l^*} \left(1 + \frac{l^*}{N}\right)^{N + l^*} e^{-(N + l^*)}}
\\
&= \frac{N^N}{N^{N + l^*}} \left[ \frac{e^{l^*}}{\left(1 + \frac{l^*}{N}\right)^{N + l^*}} \right]
\\
&= N^{-l^*} \left[ \frac{e^{l^*}}{\left(1 + \frac{l^*}{N}\right)^{N + l^*}} \right] .
\end{align}
The factor of $n$ as in \ref{eqn:volume_M} comes from $n$ replicas.

\vspace{2mm}

\textbf{Lemma 3.} Define $\mathcal{S}_{\text{prior}}$ as in \ref{app:complex}. It follows that
\begin{align}
\mathcal{S}_{\text{prior}}(\alpha_c) = (1+\rho)\log(1+\rho) - \rho\log\rho .    
\end{align}

\vspace{2mm}

\textit{Proof.} We defined 
\begin{align}
\mathcal{S}_{\text{prior}} = \left. \frac{\partial}{\partial n} \log c_{\mu^*}(N) \right|_{n=0}
\end{align} 
We can note, using strategies in section \ref{sec:analysis}, 
\begin{align}
e^{N G_0(Q)} = \det(I - Q)^{-N}.
\end{align}
In the Schur case, using the identity of \ref{eqn:cauchy_littlewood},
\begin{align}
\det(I - X)^{-N} = \sum_{\mu} s_\mu(I_N) s_\mu(X),
\end{align}
so explicitly via coefficient matching of \ref{eqn:coefficient_matching}
\begin{align}
c_\mu(N) = s_\mu(I_N).
\end{align}
This follows since we also know the decomposition $\exp(c_{N,P} G(A)) = \sum_{\lambda} c_\lambda(P) s_\lambda(A)$, giving us our coefficient-matching procedure. We use for our given partition $\mu = (l^n), \quad l = \rho N$. Therefore, via evaluation of Schur polynomials \cite{macdonald1995symmetric}
\begin{align}
\left. \frac{\partial}{\partial n} \log s_{(l^n)}(I_N) \right|_{n=0} = N \left[ (1+\rho)\log(1+\rho) - \rho\log\rho \right] + o(N).
\end{align}
Therefore
\begin{align}
\frac{1}{N} S_{\text{prior}} \longrightarrow (1+\rho)\log(1+\rho) - \rho\log\rho.
\end{align}

$ \square $

\section{Additional discussion}
\label{app:discussion}

We highlight some proof strategies relevant to the claim as in section \ref{app:real_phase}. Let us consider $\int_{-\infty}^{\infty} \delta(x-y)dx = 1$. Multiplying by a generalized function, we must have
\begin{align}
\int_{-\infty}^{\infty} dy f(y)  & = \int_{-\infty}^{\infty} dy \int_{-\infty}^{\infty} dx \delta(x-y) f(y)  \\
& = \int_{-\infty}^{\infty} dy \int_{-\infty}^{\infty} dx \delta(x-y) f(x) ,
\end{align}
since the Dirac measure forces a relationship on $x,y$. This identity is close to that used in \ref{app:real_phase}, but in general, the scenario in \ref{app:real_phase} is more nontrivial due to use of $R(\Theta)$. This is why the "constant on the fibers of $\Theta \rightarrow R(\Theta)$" argument is necessary. Now, let us consider
\begin{align}
\int_{-\infty}^{\infty} dy g(x,y) = 1 .
\end{align}
The above is irrelevant to a sifting property. However, it is true via a nascent Dirac delta that 
\begin{align}
\lim_{\epsilon \to 0^+} \int_{-\infty}^{\infty}dy g_\epsilon(x-y) f(y)  = f(x) ,
\end{align}
which is what the claim as in \ref{app:real_phase} conveys.

\section{Gardner volumes and equivalences for different $\Theta$}

\textit{\textbf{Theorem 2.} Let $\mathbb{S}_N = \{\Theta \in \mathbb{C}^N \mid \|\Theta\|_2^2 = N\}$ and define the uniform spherical measure $d\mu_{S}(\Theta) \propto \delta(\|\Theta\|_2^2 - N) d\Theta$. Let $d\mu_{G}(\Theta) = \pi^{-N} \exp(-\Theta^\dagger \Theta) d\Theta$ be the complex Gaussian measure under the flat metric $h_{j\overline{k}} = \delta_{j\overline{k}}$. Suppose for a single pattern $z^\mu \sim \mathcal{CN}(0, I_N)$ , and let $\mathbb{X}(\Theta, z^\mu) = H(\text{Re}(\Psi^\mu) - \kappa)$. Assume that $Q_{aa}^* = 1 + o(1)$ in the $d\mu_G$ case, which is necessary because otherwise the Gaussian model would concentrate at a different radius than the spherical model. In the thermodynamic limit $N, P \to \infty$ with $P/N = \alpha$, the quenched free energy densities evaluate to the same
\begin{align} \lim_{N \to \infty} f_{S} = \lim_{N \to \infty} f_{G} . \end{align}}

\textit{Proof.} The proof has similarities to Lemma 2 in Appendix \ref{app:proofs}. We start with the replicated fractional volume $\langle V^n \rangle$, which we were working with in Appendices \ref{app:real_phase}, \ref{app:complex}. For both measures, we introduce $Q \in \mathbb{S}_n$ using the Dirac measure identity
\begin{align}
\label{eqn:dirac_thm2}
1 = \int \mathcal{D}Q  \delta_{\mathcal{H}_n} \left(Q - \frac{1}{N}\Theta^\dagger \Theta\right) .
\end{align}
For the Gaussian measure, the replicated volume is
\begin{align}
\langle V_{G}^n \rangle = \int \mathcal{D}Q \int \prod_{a=1}^n d\mu_{G}(\Theta^a) \delta_{\mathcal{H}_n} \left(Q - \frac{1}{N}\Theta^\dagger \Theta\right)  \left( \mathbb{E}_Z \left[ \prod_{a=1}^n \mathbb{X}(\Theta^a, Z) \right] \right)^P .
\end{align}
The expectation over the data $Z$ depends only on $Q$ (this argument is reminiscent of that we saw in \ref{app:discussion}) so we write the expected/pattern term as $\exp(n \alpha N \mathcal{G}_c(Q))$. Similarly to \ref{app:discussion}, we rewrite
\begin{align}
\langle V_{G}^n \rangle = \int \mathcal{D}Q \exp\Big(\log M_G(Q, n) + n \alpha N \underbrace{ \mathcal{G}_c(Q) }_{\text{pattern term}} \Big) .
\end{align}
The interior integral is absorbed into the definition of $M_G$. With respect to $d\mu_{G}(\Theta)$, the diagonal entries $Q_{aa}$, for example as in the ansatz in \ref{app:real_phase} have the scaling $\text{Var}(Q_{aa}) = \mathcal{O}(1/N)$. We remark this can partially be noted in \ref{eqn:dirac_thm2}. Therefore, as $N \to \infty$, we recover
\begin{align}
q_d = \frac{1}{N}\|\Theta\|_2^2 \xrightarrow[N\to\infty]{\text{a.s.}} 1, \quad \text{or more aptly, due to assumption} \quad  q_d = 1 + o(1) . 
\end{align}
With respect to $d\mu_{S}(\Theta)$, this same condition is enforced at finite $N$ by the spherical constraint. Thus, $Q_{aa}=1$ from the start. Note that this is consistent with the replica ansatz. The spherical prior volume $M_S(Q,n)$ is therefore the Gaussian case under the restriction $Q_{aa}=1$. Now, we apply Laplace's method as $N \to \infty$. In the Gaussian case, the integral over $Q$ is dominated by a saddle point $Q^*$ whose diagonal entries satisfy $Q^*_{aa}\to1$. 

\vspace{2mm}

Both models have the same off-diagonal structure, which we saw in \ref{app:real_phase}, and the Gaussian saddle approaches the same diagonal as in the spherical case. Hence, the two extremal actions agree
\begin{align}
\text{extr}_{Q} \Big[ \log M_G(Q, n) + n \alpha N \mathcal{G}_c(Q) \Big] = \text{extr}_{Q \mid Q_{aa}=1} \Big[ \log M_S(Q, n) + n \alpha N \mathcal{G}_c(Q) \Big] .
\end{align}
Since $f$ is characterized by the above and since they agree, we get the result. In particular, recall $f$ follows the form
\begin{align}
f = -\lim_{N \to \infty} \frac{1}{N} \lim_{n \to 0} \frac{\log \langle V^n \rangle}{n} ,
\end{align}
so the two cases agree with
\begin{align}
f_{S} = f_{G}.
\end{align}

$\square$

\vspace{2mm}

\textit{Remark.} To make the relation between the Gaussian and spherical measures more rigorous, let
\begin{align}
\Theta &\sim \mathcal{CN}(0,I_N),\qquad q=\frac{\|\Theta\|_2^2}{N}, \qquad \Theta=\sqrt{q}\widehat{\Theta},
\end{align}
where $\|\widehat{\Theta}\|_2^2=N$. We can write
\begin{align}
\label{eqn:p_N}
d\mu_G(\Theta) &= p_N(q) dq d\mu_S(\widehat{\Theta}), \qquad p_N(q) = \frac{N^N}{\Gamma(N)}q^{N-1}e^{-Nq}, 
\end{align}
which follows from polar coordinates \cite{kaptanoğlu2017realcomplexintegralsspheres}. By Stirling's formula on \ref{eqn:p_N},
\begin{align}
\frac{1}{N}\log p_N(q) &= -\big(q-1-\log q\big)+o(1),
\end{align}
up to terms independent of $q$. Therefore, we can note $I(q) = q-1-\log q\geq 0, I(q)=0\iff q=1$.
We can also note $ p_N(q) \asymp e^{-NI(q)}$.  For $n$ replicas, write $\Theta^a=\sqrt{q_a}\widehat{\Theta}^a$. Then $ Q_{ab} = \frac{1}{N}(\Theta^a)^\dagger\Theta^b = \sqrt{q_aq_b}\widehat Q_{ab},  \widehat Q_{ab} = \frac{1}{N}(\widehat{\Theta}^a)^\dagger\widehat{\Theta}^b$,
with $\widehat Q_{aa}=1$. Under the assumption $q_a=Q_{aa}^*=1+o(1)$, so $\sqrt{q_a q_b} = 1 + o(1)$. It follows that
\begin{align}
Q_{ab} &= \widehat Q_{ab}+o(1).
\end{align}

\section{Additional figures}

\begin{figure}[H]
  \centering
  \includegraphics[width=0.5\linewidth]{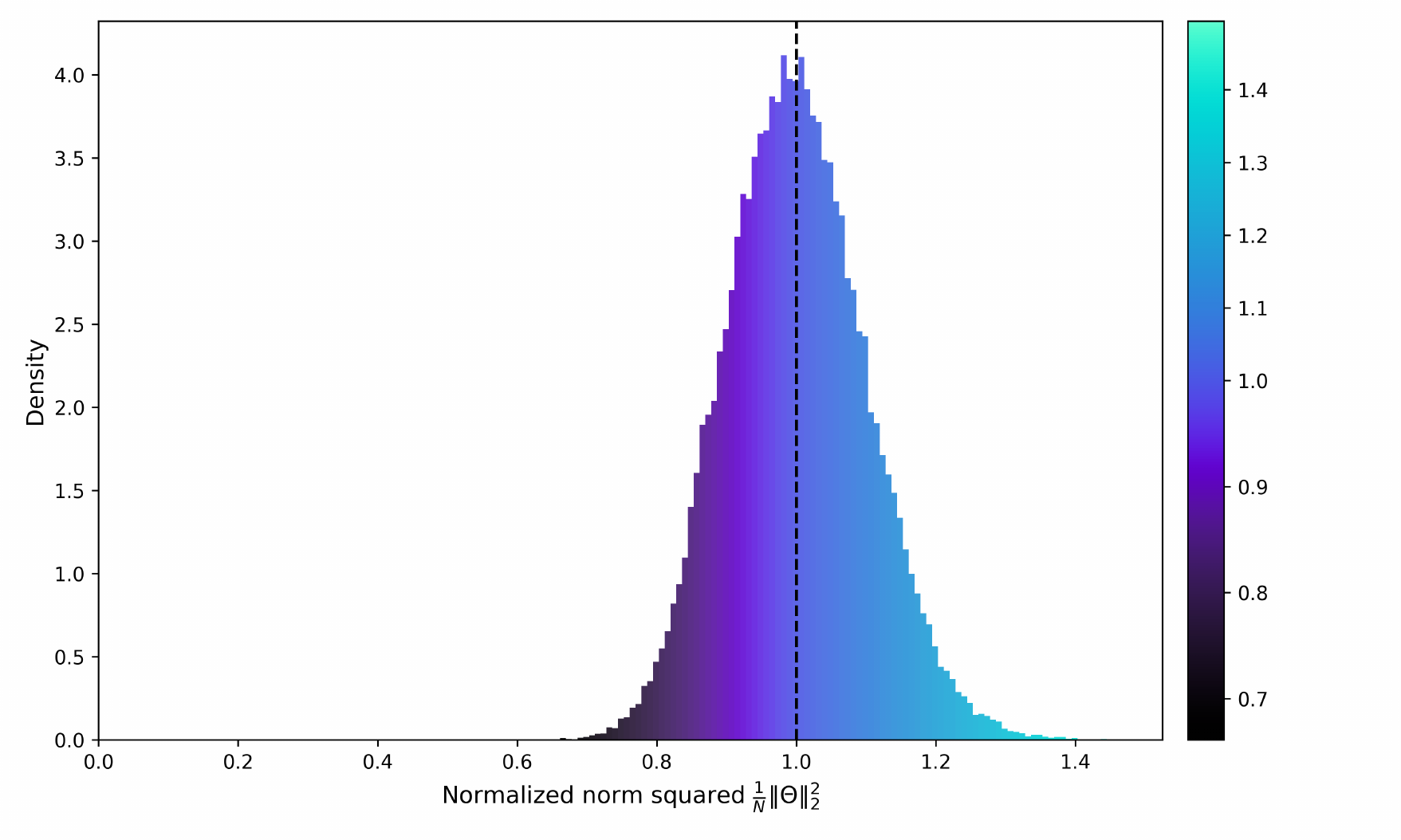}
  \vspace{-2mm}
  \caption{We create an empirical density of $\frac{1}{N} \| \Theta \|_2^2$ across samples, and the extremum occurs at 1. This plot was created with real data with dimension $N=100$ and 50,000 samples.}
  \label{fig:large_sample_density_100}
\end{figure}

\begin{figure}[H]
  \centering
  \includegraphics[width=0.5\linewidth]{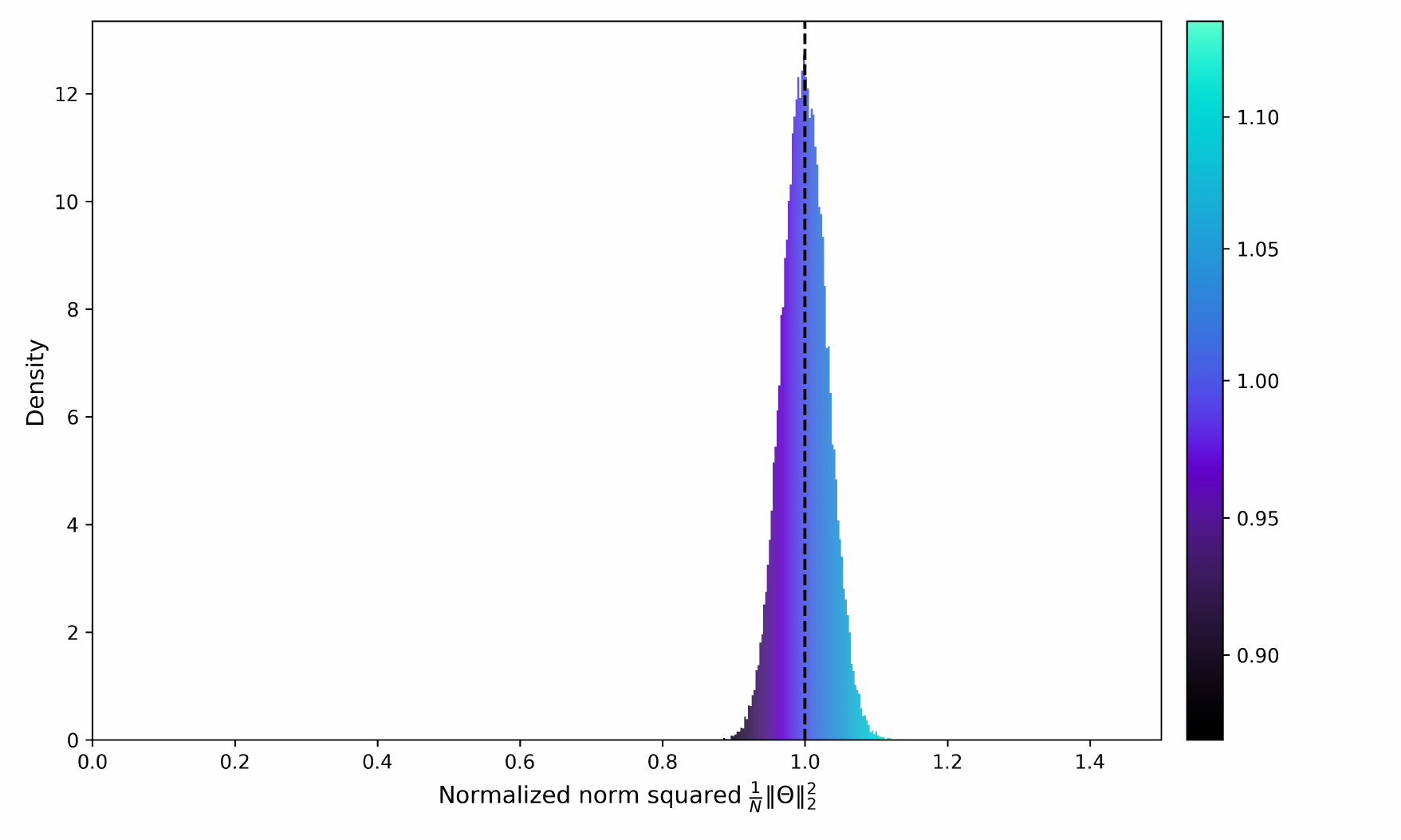}
  \vspace{-2mm}
  \caption{We plot \ref{fig:large_sample_density_100} with $N=1,000$, which demonstrates $\frac{1}{N}\|\Theta\|_2^2
\xrightarrow[N\to\infty]{\text{a.s.}} 1$.}
  \label{fig:large_sample_density_1000}
\end{figure}

\end{document}